\documentclass[runningheads]{llncs}

\usepackage{eccv}

\usepackage{eccvabbrv}

\usepackage{graphicx}
\usepackage{booktabs}
\usepackage{multirow}
\usepackage[symbol]{footmisc}
\usepackage{enumitem}

\usepackage[accsupp]{axessibility}  

\usepackage{hyperref}

\usepackage{orcidlink}

\begin{document}

\title{PanoFree: Tuning-Free Holistic Multi-view Image Generation with Cross-view Self-Guidance} 

\titlerunning{PanoFree}

\author{Aoming Liu\inst{1,2}\thanks{Work partly completed during Aoming's internship at OPPO US Research Center.}\orcidlink{0009-0007-2990-9671}  \and
Zhong Li\inst{1}\thanks{Corresponding Authors.} \orcidlink{0000-0002-7416-1216} \and
Zhang Chen\inst{1}\textsuperscript{\textdagger}\orcidlink{0000-0001-8582-1024} \and
Nannan Li\inst{2}\orcidlink{0000-0002-1545-019X} \and
Yi Xu\inst{1}\orcidlink{0000-0003-2126-6054} \and
Bryan A.\ Plummer\inst{2}\orcidlink{0000-0002-7074-3219}}

\authorrunning{A.~Liu et al.}

\institute{OPPO US Research Center, Palo Alto, CA 94303, USA \and
Boston University, Boston, MA 02215, USA}


\maketitle
\vspace{-3mm}
\centerline {\textcolor{magenta}{\href{https://panofree.github.io/}{https://panofree.github.io/}}}
\begin{abstract}
  Immersive scene generation, notably panorama creation, benefits significantly from the adaptation of large pre-trained text-to-image (T2I) models for multi-view image generation. Due to the high cost of acquiring multi-view images, tuning-free generation is preferred. However, existing methods are either limited to simple correspondences or require extensive fine-tuning to capture complex ones. We present PanoFree, a novel method for tuning-free multi-view image generation that supports an extensive array of correspondences. PanoFree sequentially generates multi-view images using iterative warping and inpainting, addressing the key issues of inconsistency and artifacts from error accumulation without the need for fine-tuning. It improves error accumulation by enhancing cross-view awareness and refines the warping and inpainting processes via cross-view guidance, risky area estimation and erasing, and symmetric bidirectional guided generation for loop closure, alongside guidance-based semantic and density control for scene structure preservation. In experiments on Planar, 360°, and Full Spherical Panoramas, PanoFree demonstrates significant error reduction, improves global consistency, and boosts image quality without extra fine-tuning. Compared to existing methods, PanoFree is up to $5$x more efficient in time and $3$x more efficient in GPU memory usage, and maintains superior diversity of results (2x better in our user study). PanoFree offers a viable alternative to costly fine-tuning or the use of additional pre-trained models. Project website at \textcolor{magenta}{\href{https://panofree.github.io/}{here}}.
\end{abstract}

\section{Introduction}
\label{sec:intro}

\begin{figure*}
  \centering
    \includegraphics[width=0.99\linewidth]{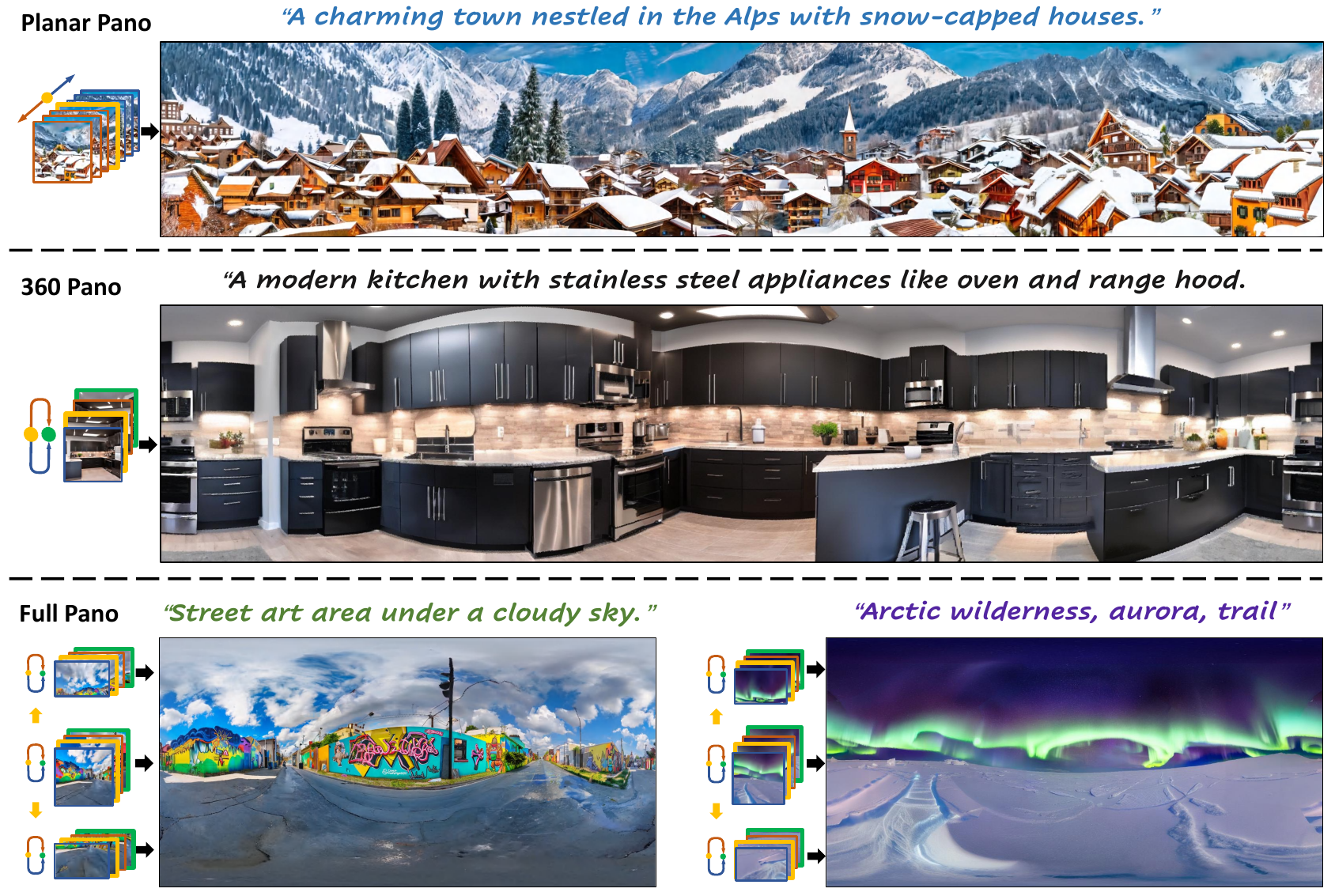}
  \vspace{-2mm}
  \caption{PanoFree can generate multi-view images according to different types of correspondences without fine-tuning, and a natural application is tuning-free generation for different types of panoramas.
We demonstrate this by generating three commonly used panoramas. Top: Planar Panorama; Middle: 360 Panorama; Bottom: Full Panorama.}
  \label{fig:pano_gen_visual}
    \vspace{-6.5mm}
\end{figure*}


Text-to-image (T2I) generation over multiple views for immersive scenes, like panorama generation, is a challenging task requiring coherence and diversity among many generated images (\eg,~\cite{brock2018large, karras2021alias, goodfellow2020generative,van2017neural, razavi2019generating, esser2021taming,ramesh2022dalle2, saharia2022imagen, ho2022classifier, ho2022cascaded}).  Early work using GANs or VAEs (\eg,~\cite{chen2023text2tex, fridman2023scenescape,Teterwak2019BoundlessGA,Lin2019COCOGANGB,Lin2021InfinityGANTI,Cheng2021InOutDI,Wang2022StyleLightHP,Chen2022Text2Light}) have been replaced recently with diffusion-based models (\eg,~\cite{Zhang2023DiffCollagePG, Li2023PanoGenTP, Wang2023Customizing3P,Feng2023Diffusion360S3,bar2023multidiffusion,Tang2023MVDiffusionEH,lee2024syncdiffusion,fang2023ctrl, voynov2023anylens, Wu2023PanoDiffusion3P}), often leveraging Stable Diffusion~\cite{rombach2022StableDiffusion}.
State-of-the-art panorama generation methods use Joint Diffusion (\eg,~\cite{bar2023multidiffusion, lee2024syncdiffusion, Tang2023MVDiffusionEH}), where parallel diffusion processes to generate multi-view images and enhancing global consistency by fusing latent or attention features based on cross-view correspondences.  However, we find these methods can only generate some types of panoramas, limiting their scope (\eg,~\cite{bar2023multidiffusion,lee2024syncdiffusion}), or require fine-tuning using expensive panorama datasets (\eg,~\cite{lee2024syncdiffusion,Tang2023MVDiffusionEH}).


To address these challenges, we propose PanoFree a tuning-free multi-view image generation method using iterative warping and inpainting of perspective images to support diverse correspondences with low costs (see \cref{fig:pano_gen_visual} for example generations).  Iterative warping and inpainting of perspective images provide a means to the diverse multi-view correspondences required in panorama generation without additional fine-tuning~\cite{hollein2023text2room, chen2023text2tex, fridman2023scenescape}.  However, recent work has overlooked these benefits due to accumulated errors causing suboptimal image quality~\cite{bar2023multidiffusion,lee2024syncdiffusion,Tang2023MVDiffusionEH}. 
We find that most accumulated errors from iterative warping can be attributed to the deficient conditions during generation. Specifically, conditioning solely on the previous image narrows cross-view awareness, leading to inconsistencies. Warping and inpainting can also propagate noise, \eg, truncated objects or jagged edges. Additionally, the given conditions may be incomplete to meet specific requirements, such as ensuring 360-degree consistency for loop closure and maintaining correct spatial relationships for realistic scenes. 

To address these issues, PanoFree expand cross-view awareness by conditioning the current view on multiple views with guided image synthesis techniques such as SDEdit~\cite{Meng2021SDEditGI}. Then, PanoFree estimates and erases the risky areas, regions likely containing noise, to reduce the noise introduced by warping and inpainting. In addition, PanoFree adopts a bidirectional generation path with a symmetrical conditioning strategy for loop closure. Lastly, PanoFree further utilizes pseudo global guidance with region-specific semantic and density control to make scene structure more reasonable.


We evaluate PanoFree on three text-to-panorama generation tasks: Planar, $360^{\circ}$, and Full Spherical Panoramas. PanoFree effectively alleviating accumulated errors in sequential generation, and significantly improves image quality and global consistency (\eg 31.6\% better in FID). This enables PanoFree to have better (or at least comparable) results to the state-of-the-art~\cite{bar2023multidiffusion,lee2024syncdiffusion,Tang2023MVDiffusionEH}, despite these methods either having narrower applications or requiring fine-tuning datasets. Specifically, PanoFree is up to 5x more efficient in time and 3x more efficient in GPU memory usage, and maintains superior diversity of results (2x better in our user study). Lastly, PanoFree is also highly flexible, enabling it to plug-in-and-play with various pre-trained T2I models and adapters.

Our contributions can be summarized as follows:
\begin{itemize}[nosep,leftmargin=*]
\item We introduce PanoFree, a tuning-free multi-view image generation method applicable for various correspondences and pre-trained T2I models. Thus PanoFree can greatly reduce the data and fine-tuning costs for immersive scene generation tasks such as text-to-panorama generation. 
\item We provide a in-depth perspective of accumulated errors and identify the deficient conditions as the main causes. We further  effectively rectify deficient conditions and alleviate accumulated errors with the cross-view guidance as well as risky area estimation and erasing in PanoFree.
\item As far as we know, PanoFree is the first to achieve feasible tuning-free generation for $360^{\circ}$ Panoramas and Full Spherical Panoramas.
\end{itemize}

\section{Related Work}
\noindent \textbf{Diffusion Models} \cite{sohl2015DMfirst,song2019NCSN,ho2020DDPM,Song2020ScoreBasedGM, song2019generative, song2020improved, karras2022elucidating} are a popular framework for generative models. 
Early work required a long trajectory for sampling  to produce high-quality samples~\cite{dhariwal2021guided-diffusion,Song2020ScoreBasedGM}, before being sped up with advanced sampling techniques that also preserved generation quality~\cite{song2020ddim,lu2022dpm-solver,karras2022edm,liu2023instaflow}.  Latent Diffusion Models (LDMs)~\cite{rombach2022StableDiffusion,peebles2023scalable} made these models more efficient by training in the latent space. 

\noindent \textbf{T2I Diffusion and Panorama Generation.} Diffusion models are widely adopted for text-to-image (T2I) generation~\cite{ramesh2022dalle2,nichol2021glide,rombach2022StableDiffusion,saharia2022imagen}. 
Many downstream tasks used large pre-trained
T2I diffusion models, like Stable Diffusion~\cite{rombach2022StableDiffusion}, to boost performance~\cite{hollein2023text2room, chen2023text2tex, fridman2023scenescape}, including panorama generation~\cite{hollein2023text2room, chen2023text2tex, fridman2023scenescape}. 
These methods have largely supplanted GAN and VAE methods~\cite{Oh2021BIPSBI, chen2023text2tex, fridman2023scenescape,Teterwak2019BoundlessGA,Lin2019COCOGANGB,Lin2021InfinityGANTI,Cheng2021InOutDI,Wang2022StyleLightHP,Chen2022Text2Light}, with most recent work in panorama generation tasks using diffusion models~\cite{Zhang2023DiffCollagePG, Li2023PanoGenTP, Wang2023Customizing3P,Feng2023Diffusion360S3,bar2023multidiffusion,Tang2023MVDiffusionEH,lee2024syncdiffusion,fang2023ctrl, voynov2023anylens, Wu2023PanoDiffusion3P}. These diffusion-based panorama generation methods use joint diffusion to fuses multiple diffusion processes through latent or attention manipulation~\cite{bar2023multidiffusion, lee2024syncdiffusion,Tang2023MVDiffusionEH}. However, they are either limited to modeling simple correspondences or require extensive fine-tuning to model complex ones. 

\noindent \textbf{Guided Image Synthesis with Diffusion Models.} It can be challenging to achieve satisfactory results solely relying on  text guidance. Therefore, some prior work~\cite{zhang2023adding, Meng2021SDEditGI, Mou2023T2IAdapterLA, Lugmayr2022RePaintIU} guide or control the generation results with reference images as fine-grained condition. ControlNet~\cite{zhang2023adding} and T2I-Adapter~\cite{Mou2023T2IAdapterLA} are the most commonly used methods to incorporate additional image conditions by adding extra image encoders, but they all require few-shot fine-tuning. SDEdit~\cite{Meng2021SDEditGI} achieves tuning-free guided image synthesis by adding noise to the guide image and then denoising it back to a real image using a pre-trained diffusion model. 

\section{Method}
\begin{figure*}[t]
  \centering
    \begin{subfigure}{0.99\linewidth}
    \centering
    \includegraphics[width=0.99\linewidth]{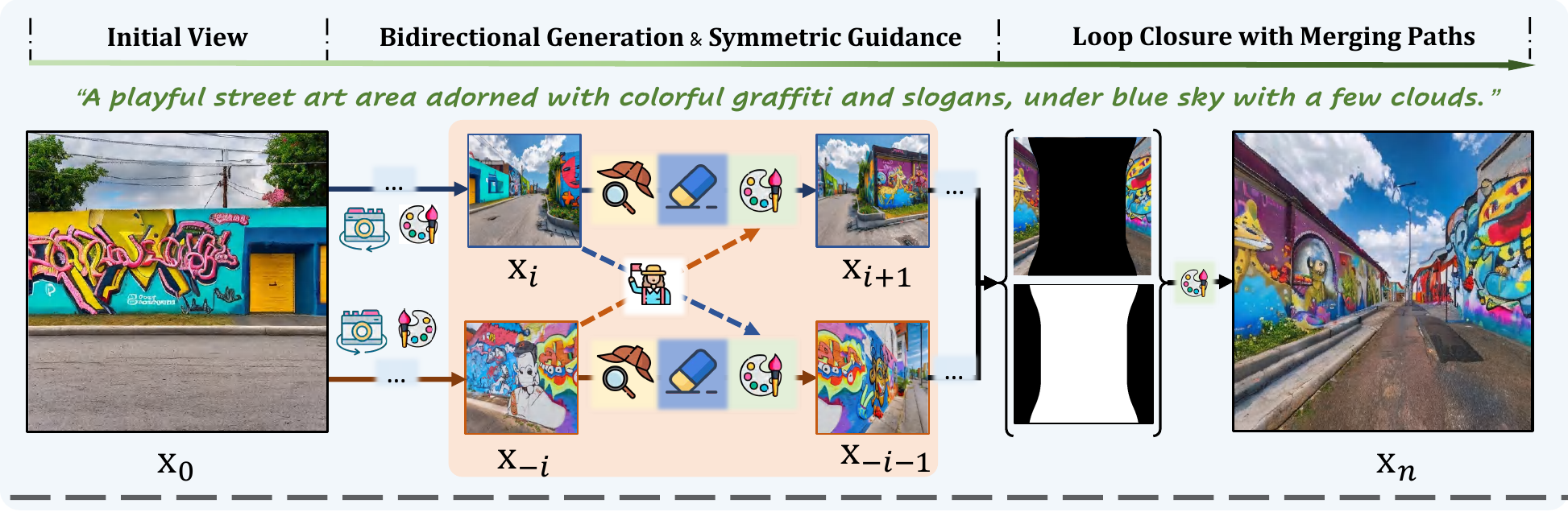}
    \label{fig:pf_framework}
    \vspace{1mm}
    \caption{Overview of the generation and guidance framework.}
  \end{subfigure}
  \vspace{1mm}
  \\
  \begin{subfigure}{0.99\linewidth}
  \centering
    \includegraphics[width=0.99\linewidth]{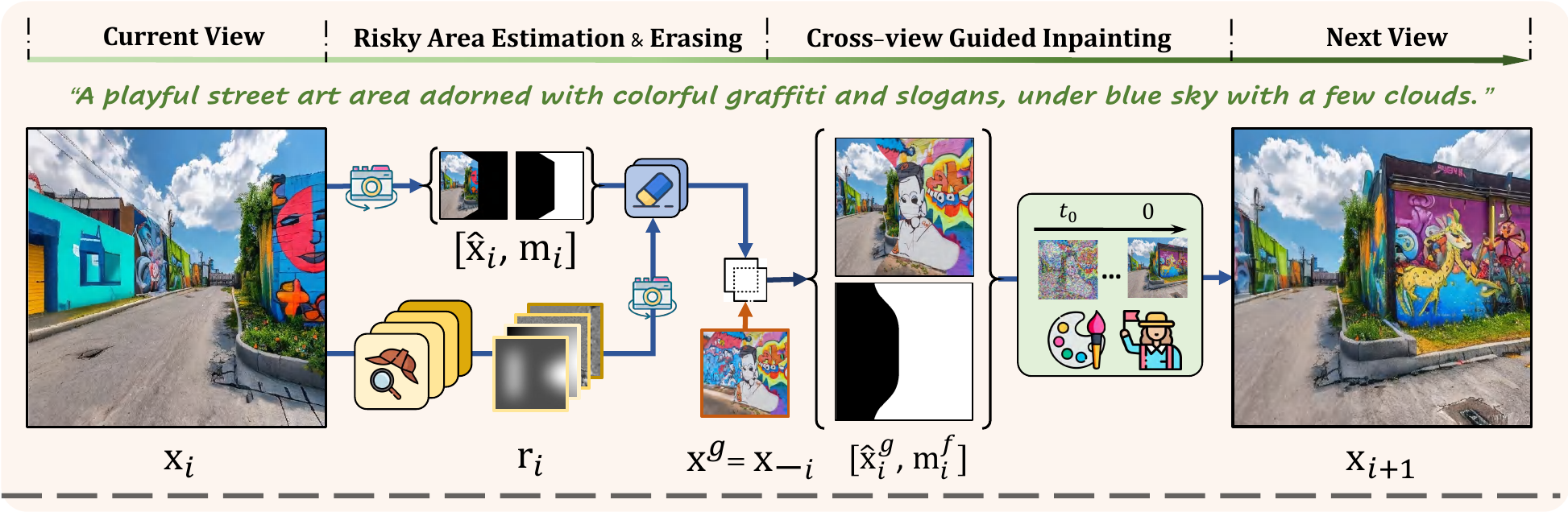}
    \label{fig:pf_step}
    \vspace{1mm}
    \caption{Illustration of a single warping and inpainting step.}
  \end{subfigure}
  \vspace{1mm}
  \\
  \includegraphics[width=0.99\linewidth]{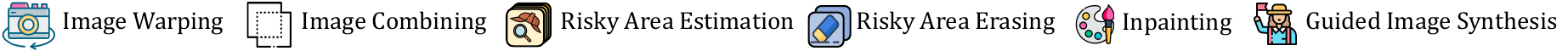}
  
  \caption{Overview of our PanoFree method, taking 360 Panorama Generation as an example. (a): At a framework level, PanoFree adopts two generation paths with opposite viewpoint translation or rotation. It enhances consistency by symmetrically selecting views from the other path as guidance to generate a new view (Sec.~\ref{sec:3.4}). Loop closure is ensured by merging these two paths. (b): In each warping and inpainting step, PanoFree reduces accumulated error by guiding the inpainting process with cross-view images (Sec.~\ref{sec:3.2}), along with estimating and erasing risky areas (Sec.~\ref{sec:3.3}). }
  \label{fig:method_overview}
    \vspace{-4mm}
\end{figure*}

PanoFree targets the text-to-panorama generation task, which takes textual descriptions as guidance to create multi-view perspective images that can be stitched into a wide-angle, high-quality panorama. PanoFree generates multi-view images through sequential warping and inpainting steps, which typically results in acclimated errors due to deficit conditions (discussed in \cref{sec:3.1}). 
 Each component of Panofree is designed to minimize the effect of these various deficiencies.  Specifically, \cref{sec:3.2} mitigates inconsistency using SDEdit-based cross-view guidance and Sec.~\ref{sec:3.3} removes artifact-inducing content by estimating and erasing risky areas. At a higher level, PanoFree employs Bidirectional Generation with Symmetric Guidance for loop closure and error reduction (Sec.~\ref{sec:3.4}). Additionally, it applies guidance-based semantic and density control for scene structure preservation (Sec.~\ref{sec:3.5}). See Fig.~\ref{fig:method_overview} for an overview of our approach.

\subsection{Deficient Conditions behind Accumulated Errors}
\label{sec:3.1}

In this section, we discuss the causes behind the deficit conditions in the iterative warping and inpainting process and reformulate the problem as conditional generation.  
Given the text prompt $c_{t}$, $i$-th view's image $\mathbf{x}_{i}$, warping function $\mathcal{W}$, transformation matrix of the projection from $i$-th view to the $(i+1)$-th view $\mathbf{P}^{i+1}_{i}$ , and pre-trained T2I inpainting model $\mathbf{\Phi}_{inp}$, the warping and inpainting step to generate the $(i+1)$-th view can be denoted as:
\begin{align}
\hat{\mathbf{x}}_{i}, \mathbf{m}_{i} = \mathcal{W}(\mathbf{x}_{i}, \mathbf{P}^{i+1}_{i}); \ \
\mathbf{x}_{i+1} = \mathbf{\Phi}_{inp}(\hat{\mathbf{x}}_{i}, \mathbf{m}_{i}, c_{t}), \ \ 
\end{align}
\noindent where $\hat{\mathbf{x}}_{i}$ is the image warped from $i$-th view to $(i+1)$-th view and $\mathbf{m}_{i}$ is the masking indicating the area to inpaint. And we can simplify the warping and inpainting steps in the following conditional image generation form:
\begin{align}
\mathbf{x}_{i+1} \sim \ q(\mathbf{x}|c_{t}, \mathbf{x}_{i}, \mathbf{P}^{i+1}_{i}).\ \
\label{eq:cond_inp}
\end{align}
However, during this generation process, we found conditions can become deficient. Major accumulated errors arise from three types of deficient conditions: Biased Conditions, Noisy Conditions, and Partial Conditions. See Sec.~A.2 in the supplementary for detailed error illustrations.

\noindent\textbf{Biased Conditions} is the most obvious problem. In the above step, $\mathbf{x}_{i+1}$ is solely conditioned on $\mathbf{x}_{i}$, which biases the cross-view awareness heavily to $i$-th view. If $\mathbf{x}_{i}$ has deviated from the desired global distribution in certain aspects, then $\mathbf{x}_{i+1}$ is likely to continue deviating in the same direction, resulting in significant inconsistency. We also found that slight style and content shifts often accumulate in this way, leading to significant inconsistency between distant views. 

\noindent\textbf{Noisy Conditions} mainly refer to $\mathbf{x}_{i}$ containing artifact-inducing contents. Existing artifacts in $\mathbf{x}_{i}$ could guide inpainting model to generate similar artifacts in $\mathbf{x}_{i+1}$ and propagate to every following view. Additionally, disjointed or distorted areas, jagged or sharp content, and objects truncated by edges in $\mathbf{x}_{i}$ are also highly risky to introduce artifacts in $\mathbf{x}_{i+1}$.  

\noindent\textbf{Partial Conditions} refer to the conditions not containing all the necessary information to meet specific requirements. For example, if we follow Eq.~\eqref{eq:cond_inp} on the final view, we lack information about $\mathbf{x}_0$, making it impossible to generate image coherent with $\mathbf{x}_0$ to ensure loop closure. Additionally, using a single text prompt to generate all views within a full spherical panorama may lead to hallucinations, such as cities floating in the sky or underwater mountains.



\subsection{Cross-View Guidance}
\label{sec:3.2}

To rectify the biased conditions discussed in Sec.\ref{sec:3.1} and enlarge cross-view awareness, a natural idea is to let $\mathbf{x}_{i+1}$ conditioned on more views,
\begin{align}
\mathbf{x}_{i+1} \sim \ q(\mathbf{x}|c_{t}, \mathbf{x}_{i}, \mathbf{P}^{i+1}_{i}, \mathbf{x}^{g}_{1},...,\mathbf{x}^{g}_{m})\ \
\label{eq:cond_inp1}
\end{align}
where $\mathbf{x}^{g}_{1},...,\mathbf{x}^{g}_{m}$ are selected from  $\mathbf{x}_{0},...,\mathbf{x}_{i-1}$. This naturally results in a Guided Image Synthesis task form with self-generated images as guidance. Many existing methods can be adapted to implement our design, such as ControlNet~\cite{zhang2023adding} and T2I adapter~\cite{Mou2023T2IAdapterLA}. To avoid relying on fine-tuning and reduce memory and time costs, we choose SDEdit~\cite{Meng2021SDEditGI}, a training-free guided image synthesis approach, with a single guidance image $\mathbf{x}^{g} \in [\mathbf{x}_{0},...,\mathbf{x}_{i-1}]$.

\noindent \textbf{Guided Image Synthesis using SDEdit.} Given $\mathbf{x}^{g}$ as guidance, SDEdit establishes a Gaussian distribution using $\mathbf{x}^{g}$ as the expectation and the intermediate status at time $t_{0}$ in the reverse SDE process. The desired data distribution is obtained by gradually removing noise from $\mathbf{x}^{g}(t_0)$:
\begin{align}
\mathbf{x}^{g}(t_0) \sim \mathcal{N}(\mathbf{x}^{g}, \sigma^2(t_0)\mathbf{I}); \ \ \mathbf{x} \sim \text{SDEdit}(\mathbf{x}^{g}, t_0, \mathbf{\Phi}),
\end{align}
where $\mathbf{\Phi}$ denotes a generative model. In PanoFree, we use inpainting mask $\mathbf{m}_{i}$ to paste the guidance image $\mathbf{x}^{g}$ to the blank areas in the warped image $\hat{\mathbf{x}}_{i}$ and then use SDEdit in the inpainting process:
\begin{align}
\hat{\mathbf{x}}^{g}_{i} = \mathbf{m}_{i} \cdot \mathbf{x}^{g} + (1 - \mathbf{m}_{i}) \cdot \hat{\mathbf{x}}_{i}; \ \ \mathbf{x}_{i} \sim \text{SDEdit}(\hat{\mathbf{x}}^{g}_{i}, t_0, \mathbf{\Phi}_{inp}).
\end{align}
Since we only want to use additional guidance images to rectify the biased conditions rather than replicate the guidance image, we use $t_0 \in [0.9, 1.0)$ in practice. Meanwhile, we found that different selection of generation path and guidance image results in different generation qualities, and the optimal choice may vary for different tasks. We introduce a general selection effective for various tasks in Section~\ref{sec:3.4} and provide an example of extending this technique to make scenes more realistic in specific scenarios in Section~\ref{sec:3.5}.

\subsection{Risky Area Estimation and Erasing}
\label{sec:3.3}

To rectify the noisy conditions discussed in Sec.\ref{sec:3.1} and eliminate the accumulation of artifacts, a natural idea is to detect and localize the artifact-inducing contents, and erase them. However, precise detection and localization often requires costly training. Thus, we turn to roughly estimate and erase the risky areas that are likely to contain artifact-inducing contents, based on indicators often associated with artifact: distances, color and smoothness. See Sec.~A.2 in the supplementary for examples. 

\noindent\textbf{Risk Estimation based on Distances.} We consider the distance from the center point of the initial view $\mathbf{x}_{0}$ and the distance to the edges. 
This is based on two priors: 1. The farther from the initial view, the more accumulated errors and the more likely to contain artifact-inducing contents. 2. Areas close to the edges are highly risky because truncated objects are mostly generated around the edges, and areas near the edges are often more severely distorted during warping.
We use initial risk $\mathbf{r}_{init}$ to represent the risk estimated based on the distance from the initial view, and edge risk $\mathbf{r}_{edge}$ to represent the risk estimated based on the distance from edges. They are derived from the following:
\begin{align}
\mathbf{r}_{init}(\mathbf{c}_{i}) = \mathcal{R}_{p}(\mathcal{D}_{0}(\mathbf{c}_{i})); \ \ 
\mathbf{r}_{edge}(\mathbf{c}_{i}) = \mathcal{R}_{p}(\mathcal{D}_{\mathbf{e}}(\mathbf{c}_{i})),
\end{align}
where $\mathbf{c}_{i}$ represents the pixel coordinates of $\mathbf{x}_{i}$ within the panorama coordinate system, $\mathcal{D}_{0}$ measures the distance to the center point of the initial view along the generation path, $\mathcal{D}_{\mathbf{e}}$ measures the distance to all edges $\mathbf{e}$, and $\mathcal{R}_{p}$ is a scaling function. We use weighted euclidean distance for $\mathcal{D}_{0}$, Gaussian filters for $\mathcal{D}_{\mathbf{e}}$, and min-max normalization for $\mathcal{R}_{p}$.

\noindent\textbf{Risk Estimation based on Color and Smoothness.} After generating a view, we can predict the risk based on color and smoothness. This uses two priors: 1. Artifacts are often not smooth or distinct in color. 2. Salient areas with abrupt colors or unevenness are prone to causing artifacts. Color-based risk $\mathbf{r}_{color}$ and smoothness-based risk $\mathbf{r}_{smooth}$ are estimated in similar forms:
\begin{align}
\mathbf{r}_{color}(\mathbf{x}_{i}) = \mathcal{R}_{f}(\mathcal{D}_{c}(\mathbf{x}_{i})); \ \ 
\mathbf{r}_{smooth}(\mathbf{x}_{i}) = \mathcal{R}_{f}(\mathcal{D}_{s}(\mathbf{x}_{i})),
\end{align}
where $\mathcal{D}_{c}$ and $\mathcal{D}_{s}$ measures the abruptness of each pixel based on color and smoothness. When implementing them, we choose pixels with the same vertical coordinates across views, and calculate the ``distances'' of each pixel to the mean color and color gradient. Within $\mathcal{R}_{f}$, we applied Gaussian filtering after min-max normalization, as those estimated risks are usually noisy.

\noindent\textbf{Erasing with Estimated Risks.} With the estimated risks, we can erase the risky areas on the image warped to next view and the inpainting mask. Assume that we get inpainting mask for current view $\mathbf{m}_{i}$ and risks for previous view $\mathbf{r}_{i-1} = [\mathbf{r}^{i}_{i-1}, \mathbf{r}^{e}_{i-1},\mathbf{r}^{c}_{i-1}, \mathbf{r}^{s}_{i-1}]$. The risks are combined linearly and new inpainting mask for current view can be obtained with:
\begin{align}
\mathbf{m}^{r}_{i} = \mathcal{M}_{r}(\mathbf{m}_{i}, \mathcal{W}(\mathbf{r}_{i-1} \cdot \mathbf{w}, \mathbf{P}_{i-1}^{i})).
\end{align}
$\mathcal{M}_{r}$ is the risk-based remasking function, and $\mathbf{w}$ are user defined combination weights. We define $\mathcal{M}_{r}$ as thresholding the risk within the warped area.

\noindent\textbf{Smoothing and Anti-aliasing.} We note that the inpainting mask from risk-based erasing may not be smooth. Additionally, sharp and jagged edges on the inpainting mask can lead to artifacts. Therefore, we also employ fixed filtering $\mathcal{M}_{f}$, where Gaussian filtering and thresholding are used to smooth the mask and reduce sharp edges, while median filtering is used to reduce jagged edges. Then, we use the final inpainting mask for the combination with guidance, and the risky areas on the warped image are removed and regenerated.
\begin{align}
\mathbf{m}^{f}_{i} = \mathcal{M}_{f}(\mathbf{m}^{r}_{i});\ \ 
\hat{\mathbf{x}}^{g}_{i} = \mathbf{m}^{f}_{i} \cdot \mathbf{x}^{g} + (1 - \mathbf{m}^{f}_{i}) \cdot \hat{\mathbf{x}}_{i}.
\end{align}

\subsection{Bidirectional Generation with Symmetric Guidance}
\label{sec:3.4}

\textbf{Bidirectional Generation.} We begin by dividing a unidirectional generation path $\mathbf{x}_{0} \rightarrow \mathbf{x}_{1}... \mathbf{x}_{2n}$ into two bidirectional generation paths $\mathbf{x}_{0} \rightarrow \mathbf{x}_{1} \rightarrow ... \rightarrow \mathbf{x}_{n}$ and $\mathbf{x}_{-n} \leftarrow ... \leftarrow \mathbf{x}_{-1} \leftarrow \mathbf{x}_{0}$. Typically, we would make these two generation paths symmetric. And we found this can reduce accumulated errors because the distance to the initial view is reduced in each direction. This consistently reduces artifacts, but may not reduce style and content inconsistency, as there may be different style/content shift in the two directions. 

\noindent\textbf{Loop Closure.} To ensure loop closure, we can add a $(2n + 1)$-th view as the "merging view" to merge the 2 generation paths by warping $\mathbf{x}_{n}$ and $\mathbf{x}_{-n}$ to the $(2n + 1)$-th view and inpaint it. However, if the differences between the two paths are too large, $\mathbf{x}_{2n + 1}$ may contain image tearing, failing to ensure loop closure. This is due to the partial conditions on each path: there is no information from the other path before merging. Therefore, we rectify the partial conditions by introducing awareness of the other path.

\noindent\textbf{Symmetric Guidance.} We introduce the awareness of the other path by selecting symmetric guidance images from the other path. Specifically, when generating $\mathbf{x}_{i+1}$, we will select $\mathbf{x}_{-i}$ as the guidance image. Thus, $\mathbf{x}_{i+1}$ will get the awareness of both paths as it is conditioned on $\mathbf{x}_{i}$ and $\mathbf{x}_{-i}$:
\begin{align}
\mathbf{x}_{i+1} \sim \ q(\mathbf{x}|c_{t}, \mathbf{x}_{i}, \mathbf{x}_{-i}, \mathbf{P}^{i+1}_{i})\ \
\end{align}
We emperically found that bidirectional generation with symmetric guidance is not only effective in ensuring loop closure but also a universally applicable strategy to effectively reduce accumulated errors in various scenarios.

\subsection{Aligning with Scene Structure Prior}
\label{sec:3.5}
When generating full spherical panoramas, we divide a spherical panorama into five parts: first, we generate a 360 panorama as the central part, then we expand upwards and downwards, and finally, we generate two images centered around the top and bottom poles to close up the entire spherical surface. During the expansion and closing stages, models often fail to align with scene structure priors due to partial conditions and generate artifacts.

\noindent\textbf{Hallucination} refers to the artifacts caused by mismatches between partial conditions and scene structure priors. For example, when generating a city scene, using the same prompt during the expansion and closing stages may result in a floating city in the sky or a city underwater. The most direct solution is to input a new prompt, but this would require additional manual effort, which is not ideal. So, we attempt to rectify the partial conditions by extracting scene structure priors and applying semantic and variance control from the initial view.

\noindent\textbf{Prior Extraction.} Although the pretrained T2I model may not align a full panorama with scene structure priors, it can align a single perspective view image with them. Therefore, we extract the scene structure prior from the initial view image $\mathbf{x}_{0}$ and incorporate it into the expansion process. For example, when generating the first view image in the upward expansion $\mathbf{x}^{ue}_{0}$, we use upper $1/3$ part of the initial view image as guidance with resizing it to the size of $\mathbf{x}^{ue}_{0}$ . 

\noindent\textbf{Semantic and Variance Tuning.} When the give text prompt only describe part of the scene, we may want generated semantic contents less conditioned on the partial prompt and more conditioned on the prior images during expansion and closing. We achieve this by reducing guidance scale and widen the field of view. Meanwhile, we adjust the variance of the initial noise to avoid the color blocks caused by low guidance scale. Through experimentation, we've found that a combination of slightly high initial variance and low guidance scale can stably reduce hallucinations and color blocks during the expansion and closing stages.

\section{Experiments}
We evaluate the performance of PanoFree across three generation tasks: Planar Panorama Generation, 360 Panorama Generation, and Full Panorama Generation. However, note that we focus on planar panorama and $360^{\circ}$ panorama generation, where the comparisons are more precise and consistent. 

\noindent\textbf{Implementation details.} PanoFree is implemented using the publicly available Stable Diffusion code from Diffusers~\cite{von-platen-etal-2022-diffusers} based on the PyTorch framework. For the experiments in the main paper, we utilized the generation and inpainting models of Stable Diffusion (SD) v2.0~\cite{rombach2022StableDiffusion}. All experiments are conducted on a single NVIDIA RTX A6000 GPU. Further details and specific configurations can be found in the corresponding sections of the main paper and the supplementary.

\noindent\textbf{Evaluation metrics.} We introduce a more comprehensive set of evaluation metrics than prior work~\cite{bar2023multidiffusion, Tang2023MVDiffusionEH, lee2024syncdiffusion} covering five themes: image quality, global consistency, prompt capability, diversity, and resource consumption. 
\begin{itemize}[nosep,leftmargin=*]
  \item {\it Image Quality} is measured with Fréchet Inception Distance (FID)~\cite{Heusel2017GANsTB}, Kernel Inception Distance (KID)~\cite{Binkowski2018DemystifyingMG}, which measure fidelity and diversity. FID and KID calculated between the views randomly cropped from the panorama and reference images generated by SD with the same prompts. 

  \item {\it Global Consistency} is measured with Intra-LPIPS (IL)~\cite{Zhang2018TheUE} used by SyncDiffusion~\cite{lee2024syncdiffusion}, which is computed by cropping non-overlapping views from a panorama and computing the averaged LPIPS scores of all view pairs.
  
  \item {\it Prompt Capability} is measured via CLIP Score (CS)~\cite{Hessel2021CLIPScoreAR} by computing the text-image similarity of randomly cropped views of the panorama.
   
  \item {\it Panorama Diversity} is also measured by FID and KID. Additionally, we propose Cross-LPIPS (CS)~\cite{Zhang2018TheUE}. Cross-LPIPS is computed across 2 panoramas generated with a same text with differents random seeds. We crop non-overlapping views from each panorama, and compute the averaged LPIPS scores of all view pairs where two views come from different panoramas. 
  
  \item {\it Resource Consumption} includes time consumption, measured by the cumulative time cost of all diffusion processes to generate a single panorama, and peak GPU memory consumption, measured by the maximum GPU memory consumption during inference. 
\end{itemize}

\noindent\textbf{Evaluation Settings.} Prior work either used arbitrary prompts~\cite{bar2023multidiffusion,lee2024syncdiffusion} or only focused on a single type of scene~\cite{Tang2023MVDiffusionEH}. Instead, we consider 3 distinct scene types: indoor, street, and city scenes, and natural scenes. We obtained 100 prompts for each type from ChatGPT~\cite{radford2019language}. We use 10  random seeds per prompt for planar panorama and 360 panorama generation, and 3 different random seeds per prompt for full panorama generation (see supplementary for details). 

\noindent\textbf{User Study.} For planar panorama generation and $360^{\circ}$ panorama generation, we conducted four user studies for each task to further evaluate the global consistency, image quality, prompt compatibility, and diversity of the generated panoramas (see supplementary for details). 

\subsection{Planar Panorama Generation}
\label{sec:4.1}
Planar Panorama corresponds to the scene observed with camera translation along the focal plane in reality. This is a relatively simple task, as it only involves extending the image without considering more complex geometric changes. 
\textbf{Baselines.} We have chosen 3 tuning-free baselines for comparison, {\it Vanilla Sequential Generation (SG)}, {\it MultiDiffusion (MD)}~\cite{bar2023multidiffusion} and {\it SyncDiffusion (SYD)}~\cite{lee2024syncdiffusion}. Additional details are in the supplementary.

\begin{figure*}[t]
  \centering
  \includegraphics[width=0.99\linewidth]{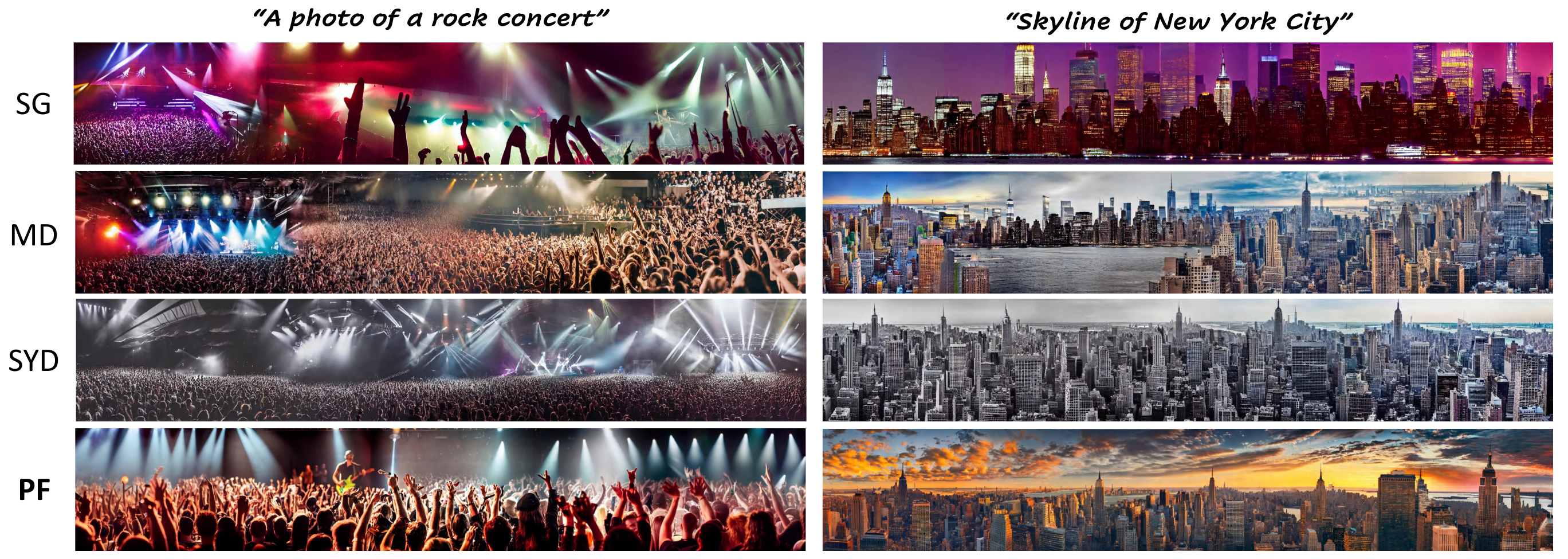}
  \vspace{-2mm}
  \caption{Planar Panorama generation results. Compared to vanilla Sequential Generation and MultiDiffusion (MD)~\cite{bar2023multidiffusion}, PanoFree achieves superior global consistency and image quality. It is also comparable to SyncDiffusion (SYD)~\cite{lee2024syncdiffusion} in these aspects.}
  \label{fig:planar_pano}
  \vspace{-4mm}
\end{figure*}

\noindent\textbf{Results.} The quantitative and qualitative evaluations are shown in Table~\ref{table:planar_evaluation} and Fig.~\ref{fig:planar_pano}, respectively. Below we compare PanoFree to each baseline. 

\begin{table}[t]
\centering
\caption{Comparison of tuning free methods for Planar Panorama generation using Stable Diffusion~\cite{rombach2022StableDiffusion}. We find PanoFree (PF) outperforms the state-of-the-art while having low computational requirements. Note that Cross-LPIPS and Intra-LPIPS are in $10^{-2}$ scale, KID is in $10^{-3}$ scale.}
\vspace{-2mm}
\resizebox{0.99\linewidth}{!}{
\begin{tabular}{cccccccc}
\toprule
 Method & Intra-LPIPS$\downarrow$ & Cross-LPIPS$\uparrow$ & FID$\downarrow$ & KID$\downarrow$ & CS$\uparrow$ & Time (s)$\downarrow$ & Memory (GB)$\downarrow$ \\
\midrule
SG  & $70.40$ & $71.03$ & $24.91$ & $4.33$ & $26.68$& $25$ & $3.2$\\
MD~\cite{bar2023multidiffusion}   & $68.48$ & $69.92$ & $21.16$ & $3.50$ & $27.89$& $95$ & $5.8$\\
SYD~\cite{lee2024syncdiffusion}  & $64.48$ & $68.07$ & $20.56$ & $3.62$ & $27.18$& $128$& $10.0$\\
\midrule
\textbf{PF (Ours)} & $65.34$ & $69.68$& $17.05$ & $3.80$& $27.21$ & $26$ & $3.2$\\
\bottomrule
\end{tabular}
}
\label{table:planar_evaluation}
\vspace{-1mm}
\end{table}

\begin{itemize}[nosep,leftmargin=*]
  \item {\it Compared with vanilla Sequential Generation,} PanoFree significantly enhances image quality and global consistency, demonstrating its effectiveness in reducing accumulated errors. Moreover, PanoFree does not compromise diversity or have a significant effect on GPU time and memory overhead. 
  \item {\it Compared with MultiDiffusion,} PanoFree has significant advantages in image quality and global consistency. Meanwhile, its time and GPU memory overhead is only 26\% and 55\% that of MultiDiffusion, respectively.
  \item {\it  Compared with SyncDiffusion}, PanoFree achieves comparable performance in global consistency and image quality. Although SyncDiffusion performs better in consistency, it requires introducing additional models for latent optimization.  This leads PanoFree's time overhead to be 20\% of SyncDiffusion and GPU memory overhead  to be 32\% of SyncDiffusion.
\end{itemize}

\noindent\textbf{The Loss of Diversity with Joint Diffusion.} When using different random seeds with the same prompt, methods using Joint Diffusion exhibit reduced diversity in their results.  In contrast, our PanoFree method can better maintain diversity (see cross-LPIPS scores in Table~\ref{table:planar_evaluation}). 
Additionally, we believe this is the source of PanoFree's gains over MultiDiffusion and SyncDiffusion in FID.

\begin{figure*}[t]
  \centering
  \includegraphics[width=0.99\linewidth]{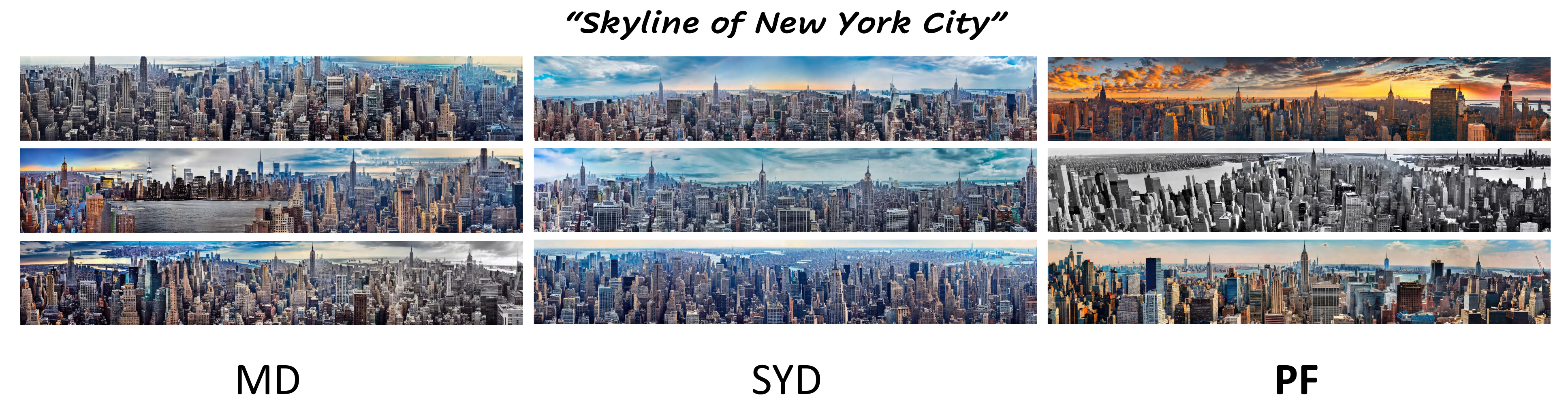}
  \vspace{-2mm}
  \caption{Diversity comparison on Planar Panorama Generation task. Each group is generated using the same text with different random seeds. Compared to MultiDiffusion (MD)~\cite{bar2023multidiffusion} and SyncDiffusion (SYD)~\cite{lee2024syncdiffusion}, PanoFree achieves superior diversity.}
  \label{fig:planar_pano_diversity}
  \vspace{-4mm}
\end{figure*}

This diversity issue becomes particularly apparent when given some underspecified prompts. Therefore, we generated a ``underspecified set'' consisting of 20 short and blurry prompts to demonstrate this issue. For each prompt, we used 20 different random seeds. 
We demonstrate the diversity differences qualitatively in Fig.~\ref{fig:planar_pano_diversity}.
Please refer to supplementary for quantitative analysis.


\noindent\textbf{User Study.}  The results in Table~\ref{table:user_study_sync} clearly show that human evaluators believe PanoFree produces more diverse panoramas and demonstrates better compatibility with prompts than SyncDiffusion~\cite{lee2024syncdiffusion}. Additionally, both methods exhibit similar levels of global consistency and image quality.

\begin{table}[t]
\centering
\caption{User study results of Planar Panorama Generation. 15 questions are used for each evaluation item and answered by 5 Amazon MTurk workers.}
\vspace{-2mm}
\begin{tabular}{ccccc}
\toprule
  & Consistency (\%)  & Quality (\%) & Prompt Compatibility (\%) & Diversity (\%)\\
\midrule
SYD~\cite{lee2024syncdiffusion}  & $52.7$ & $46.0$ & $41.3$ & $35.3$\\
\textbf{PF (ours)} & $47.3$ & $54.0$& $58.7$ & $64.7$\\
\bottomrule
\end{tabular}

\label{table:user_study_sync}
\vspace{-2.8mm}
\end{table}

\subsection{360 Panorama Generation} 
\label{sec:4.2}
Due to the distortion caused by equirectangular projection, generating 360-degree panoramas is more challenging than planar panorama generation. Vanilla sequential generation tends to produce many artifacts, significantly decreasing image quality. Moreover, MultiDiffusion~\cite{bar2023multidiffusion} and SyncDiffusion~\cite{lee2024syncdiffusion} cannot be directly used for generating 360-degree panoramas. As far as we know, PanoFree is the first implementation of training-free 360-degree panorama generation.

\noindent\textbf{Baselines.} We used 2 baselines for comparison: {\it Vanilla Sequential Generation (SG)} and {\it MVDiffusion (MVD)}. Additional details are in the supplementary.
\begin{table}[t]
\caption{Comparison of  $360^{\circ}$ Panorama generation methods using Stable Diffusion~\cite{rombach2022StableDiffusion}. We find PanoFree (PF) still outperforms the state-of-the-art while having low computational requirements. Note that Cross-LPIPS and Intra-LPIPS are in $10^{-2}$ scale, KID is in $10^{-3}$ scale.}
\vspace{-2mm}
\centering
\resizebox{0.99\linewidth}{!}{
\begin{tabular}{cccccccc}
\toprule
 Method & Intra-LPIPS$\downarrow$ & Cross-LPIPS$\uparrow$ & FID$\downarrow$ & KID$\downarrow$ & CS$\uparrow$ & Time (s)$\downarrow$ & Memory (GB)$\downarrow$ \\
\midrule
SG  & $70.62$ & $73.06$ & $32.28$ & $7.90$ & $26.35$& $21$ & $3.2$\\
MVD~\cite{Tang2023MVDiffusionEH}   & $67.71$ & $70.07$ & $37.89$ & $8.76$ & $26.27$& $110$ & $6.9$\\
\midrule
\textbf{PF (ours)} & $68.62$ & $72.67$& $25.84$ & $7.48$& $26.51$ & $22$ & $3.2$\\
\bottomrule
\end{tabular}
}
\label{table:360_evaluation}
\vspace{-4mm}
\end{table}

\begin{figure*}[t]
  \centering
  \includegraphics[width=0.99\linewidth]{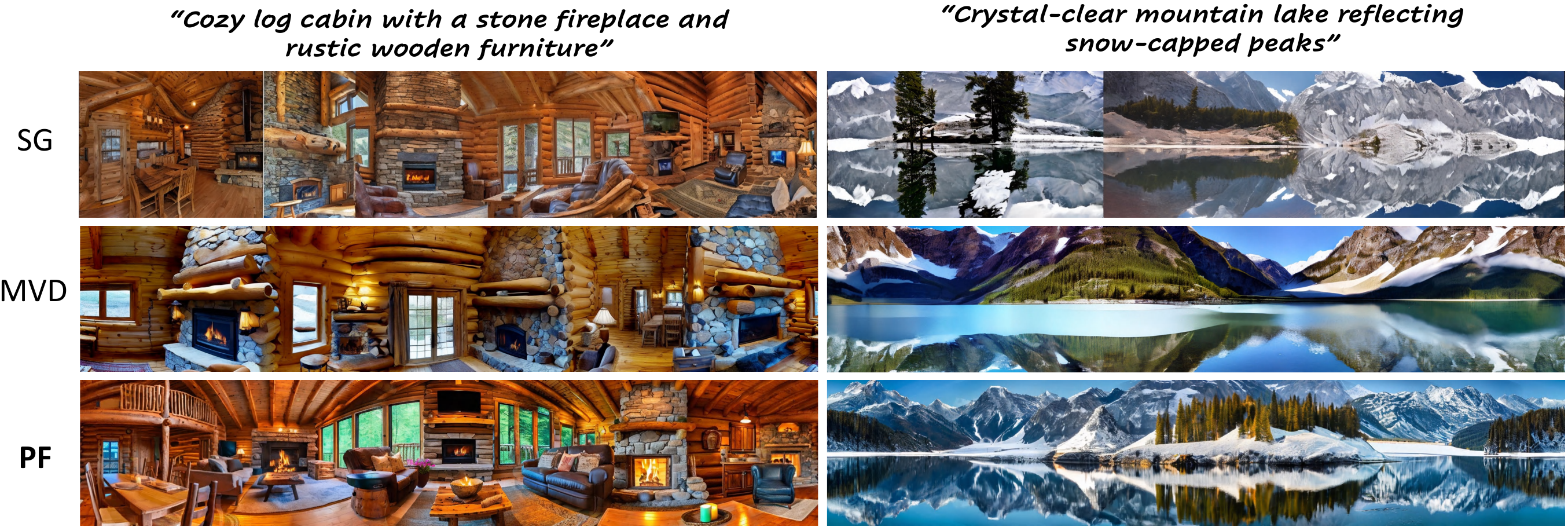}
  \vspace{-1mm}
  \caption{$360^{\circ}$ Panorama generation results. Compared to vanilla Sequential Generation (SG), PanoFree achieves superior global consistency and image quality. It is also comparable to MVDiffusion (MVD)~\cite{Tang2023MVDiffusionEH} in these aspects.}
  \label{fig:360_pano}
  \vspace{-2mm}
\end{figure*}

\begin{figure*}[t]
  \centering
  \includegraphics[width=0.99\linewidth]{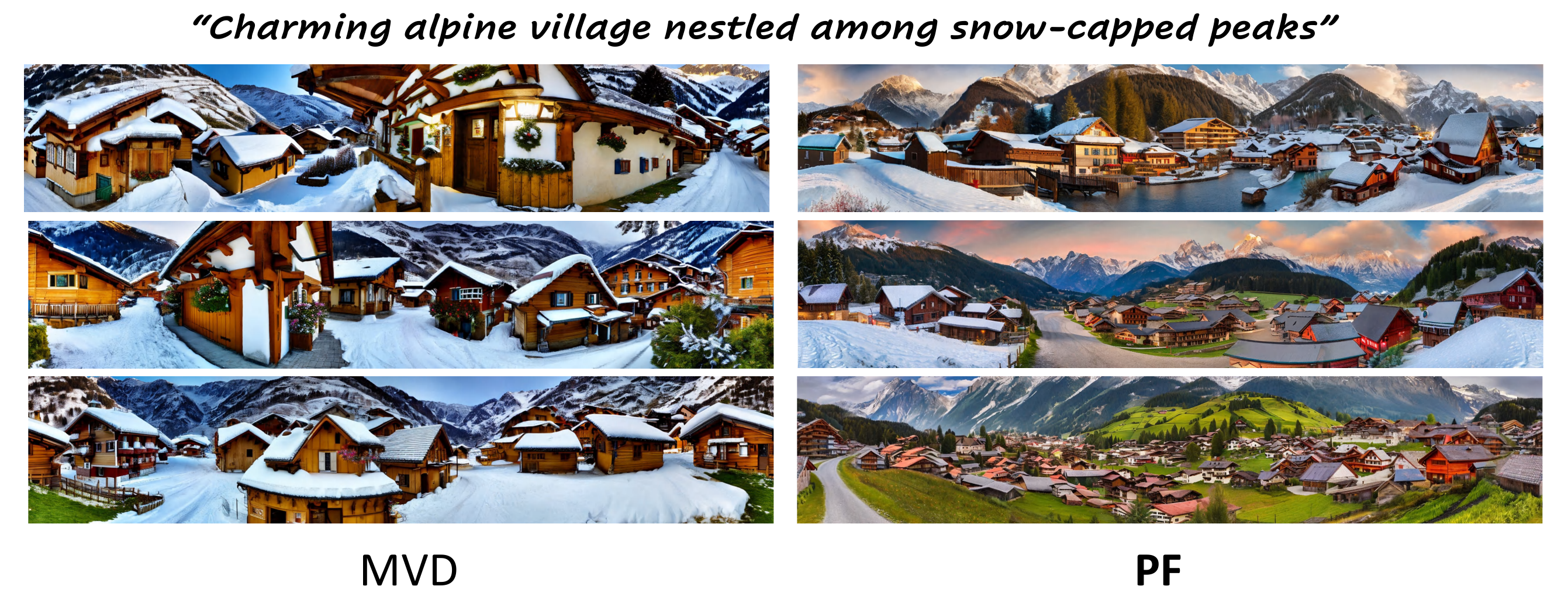}
  \vspace{-2mm}
  \caption{Diversity comparison on $360^{\circ}$ Panorama Generation task. Each group is generated using the same text with different random seeds. Compared to MVDiffusion (MVD)~\cite{Tang2023MVDiffusionEH}, PanoFree achieves superior diversity.}
  \label{fig:360_pano_diveristy}
  \vspace{-2mm}
\end{figure*}

\noindent\textbf{Results.} The quantitative and qualitative evaluations are shown in Table~\ref{table:360_evaluation} and Fig.~\ref{fig:360_pano} respectively. Below we compare PanoFree to each baseline. 
\begin{itemize}[nosep,leftmargin=*]
  \item {\it Compared with vanilla Sequential Generation,} PanoFree significantly enhances image quality and global consistency. Specifically, vanilla sequential generation creates artifacts with complex optical geometry transformations, severely impacting image quality. However, PanoFree effectively recovers image quality by estimating and erasing risky areas, minimizing artifact propagation.
  \item {\it Compared with MVDiffusion,} PanoFree achieves comparability in image quality and global consistency, yet significantly outperforms in terms of time, GPU memory overhead, and diversity. Particularly, MVDiffusion is significantly worse than PanoFree in terms of FID and KID scores, even underperforming vanilla Sequential Generation. This is partly due to the inevitable bias of MVDiffusion's generated results towards the training dataset, resulting in larger discrepancies compared to those produced by Stable Diffusion. Visually, MVDiffusion also exhibits a noticeable lack of generation diversity. As depicted in Fig.~\ref{fig:360_pano_diveristy}, given a prompt, results generated with different random seeds show minimal variation in both content and style.
\end{itemize}
\noindent\textbf{User Study.} The user study results in Table~\ref{table:user_study_mvd} show that human evaluators believe PanoFree also produces more diverse $360^{\circ}$ panoramas compared with MVDiffusion~\cite{Tang2023MVDiffusionEH}. And  PanoFree demonstrates better global consistency. Both methods exhibit similar levels of image quality and prompt comparability.

\begin{table}[t]
\centering
\caption{User study results of $360^{\circ}$ Panorama Generation. 15 questions are used for each evaluation item and answered by 5 Amazon MTurk workers.}
\begin{tabular}{ccccc}
\toprule
  & Consistency (\%)  & Quality (\%) & Prompt Compatibility (\%) & Diversity (\%)\\
\midrule
MVD~\cite{Tang2023MVDiffusionEH}  & $40.7$ & $48.7$ & $47.3$ & $33.3$\\
\textbf{PF (ours)} & $59.3$ & $51.3$& $52.6$ & $66.6$\\
\bottomrule
\end{tabular}

\label{table:user_study_mvd}
\vspace{-1mm}
\end{table}

\begin{figure*}[t]
  \centering
  \includegraphics[width=0.99\linewidth]{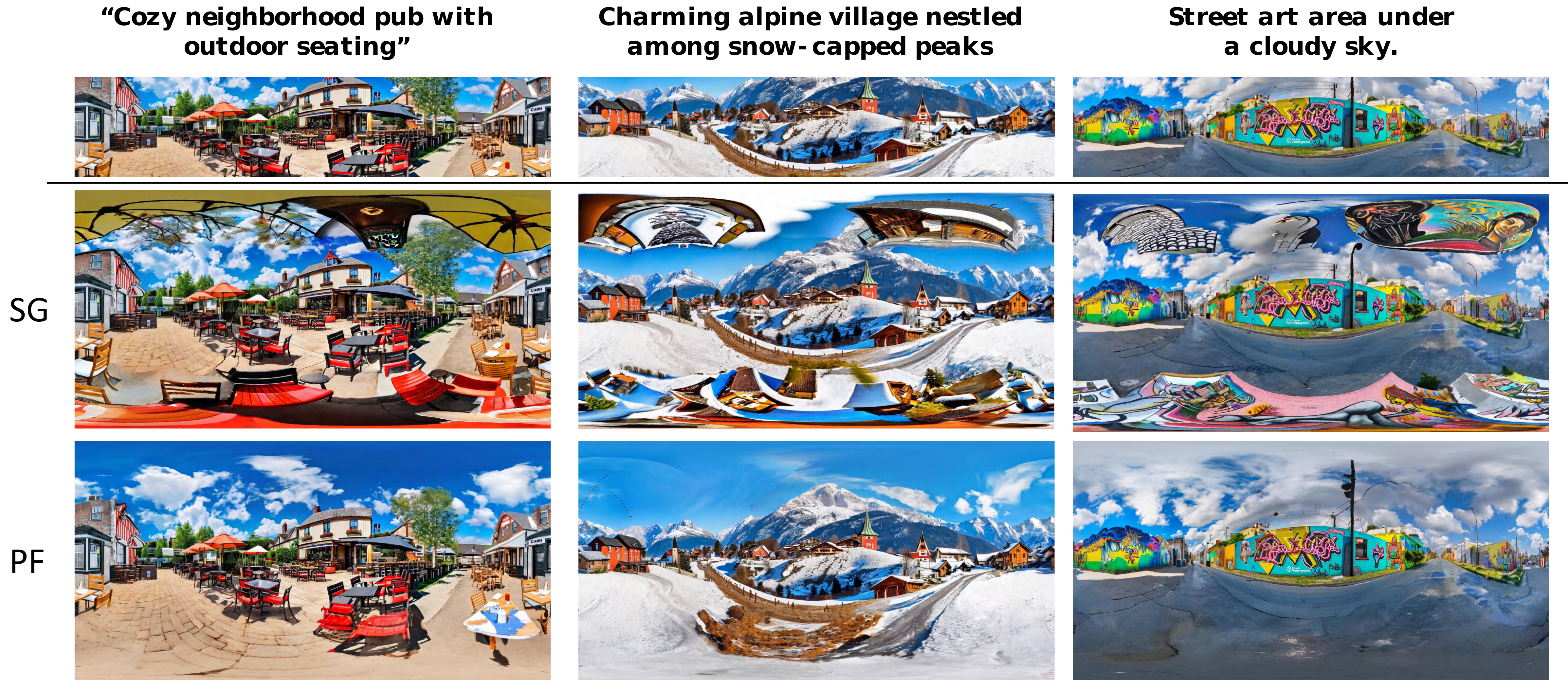}
  \vspace{-1mm}
  \caption{Full Spherical Panorama generation results. Vanilla Sequential Generation (SG) tends to generate hallucinations due to partial conditions, while PanoFree effectively mitigates this issue.}
  \label{fig:full_pano_comp}
  \vspace{-2mm}
\end{figure*}

\subsection{Full Spherical Panorama Generation}

PanoFree is also the first to achieve feasible tuning-free generation for Full Spherical Panoramas. However, it's hard to conduct meaningful comparisons due the lack of tuning-free generation baseline methods or those with  strong out-of-scope generation capabilities for Full Spherical Panoramas generation task. Thus, we conduct qualitative evaluation as well as comparison with vanilla sequential generation by showcasing generated results in Fig.~\ref{fig:full_pano_comp}. Vanilla sequential generation exhibit more artifacts as distortion increases. Additionally, partial conditioning issue mentioned in Sec.~\ref{sec:3.5} causes hallucinations. And PanoFree still could effectively reduce artifacts and hallucinations. Note that we start both methods from $360^{\circ}$ panoramas generated by PanoFree into full spherical panoramas, otherwise vanilla sequential generation will perform even worse.

\subsection{Ablation Study}
 Tab.~\ref{table:pf_ablation} contains an ablation study that sequentially integrates each PanoFree component. We evaluate consistency (Intra-LPIPS) and image quality (FID) with 30\% of the prompts from Sec.~\ref{sec:4.1} \& Sec.~\ref{sec:4.2} for both planar and $360^{\circ}$ panorama. We show that cross-view guidance provides the strongest benefit, followed by distance and edge-based risky area erasing. These components effectively reduce image tearing and visual chaos. Color and smoothness-based erasing have a smaller impact, likely due noise in these low-level features. Qualitative results are in Sec.~B of the supplementary.

\begin{table}[t]

\begin{minipage}[t]{0.395\textwidth}
    \setlength{\abovecaptionskip}{0pt}%
    \caption{Quantitative ablation of PanoFree components using 30\% of prompts from Sec.~\ref{sec:4.1} \& \ref{sec:4.2}. Intra-LPIPS ($10^{-2}$) and FID show cross-view guidance offers the most benefit, followed by distance and edge-based risky area erasing. Color and smoothness-based erasing have minimal impact.}%
    \label{table:pf_ablation}
\end{minipage}\hfill
\begin{minipage}[t]{0.60\textwidth}\vspace*{0pt}%
    \centering
    \begin{tabular}{cccc}
    \toprule
     Task & Method & Intra-LPIPS$\downarrow$ & FID$\downarrow$  \\
    \midrule
    \multirow{3}{*}{Planar} & None  & $70.84$ & $24.75$ \\
    & + CG  & $66.63$ & $20.63$ \\
    & + Dist  & $65.87$ & $18.21$ \\
    \midrule
    \multirow{4}{*}{$360^{\circ}$} & None  & $71.34$ & $33.47$ \\
     & + CG  & $69.21$ & $27.38$ \\
     & + Dist \& Edge  & $69.03$ & $26.69$ \\
     & + Color \& Smooth & $68.94$ & $26.51$ \\
    \bottomrule
    \end{tabular}
    
\end{minipage}

\vspace{-6mm}
\end{table}
 

%

\section{Conclusion}
We present PanoFree, a tuning-free multi-view image generation that supports an extensive array of correspondences. PanoFree improves error accumulation by enhancing cross-view awareness and refining the warping and inpainting processes through cross-view guidance, risky area estimation and erasing, and symmetric bidirectional guided generation for loop closure, alongside guidance-based semantic and density control for scene structure preservation. PanoFree is evaluated on various panorama types—Planar, 360°, and Full Spherical Panoramas. PanoFree demonstrates significant error reduction, improved global consistency, and image quality across different scenarios without extra fine-tuning. Compared to existing methods, PanoFree is up to $5$x more efficient in time and $3$x more efficient in GPU memory usage, and maintains superior diversity of results (2x better in our user study). Moreover, PanoFree can be extended to texture generation for 3D models. We intend to explore these possibilities in future research.

\noindent\textbf{Limitations.} A limitation of our work is that we are unable to generate scenes beyond the capability of the pre-trained T2I model. Therefore, we rely on large pre-trained T2I models to ensure the broad application scope. And when provided with text descriptions beyond the capability range of the pre-trained T2I models, the generated results may not match the text.

\bibliographystyle{splncs04}
\bibliography{main}
\appendix
\clearpage
\section{Detailed Method Illustrations}
\subsection{Panorama Generation Pipelines}
In this part, we provide detailed illustrations of PanoFree's generation processes for Planar Panoramas, 360° Panoramas, and Full Spherical Panoramas.

\begin{figure*}[h]
  \centering
  \includegraphics[width=0.99\linewidth]{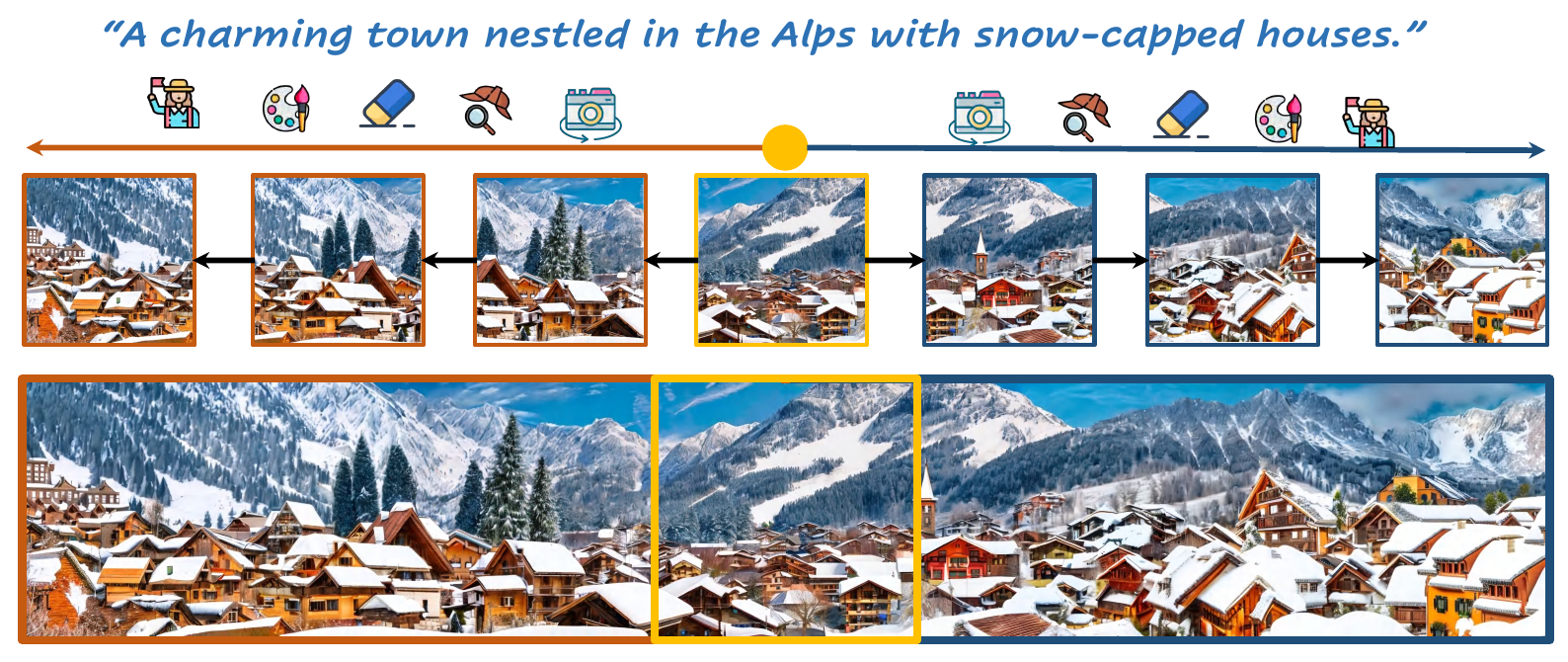}
  \vspace{-2mm}
  \caption{Detailed Illustration of Planar Panorama generation. }
  \label{fig:dil_planar}
  \vspace{-2mm}
\end{figure*}

\noindent \textbf{Planar Panorama Generation} is illustrated in Fig.~\ref{fig:dil_planar}. We use the Bidirectional Generation with Symmetric Guidance strategy to iteratively warp in two directions based on symmetric planar translation from the initial view located at the center of the planar panorama. Subsequently, we generate the image of the next view using inpainting.

\begin{figure*}[h]
  \centering
  \includegraphics[width=0.99\linewidth]{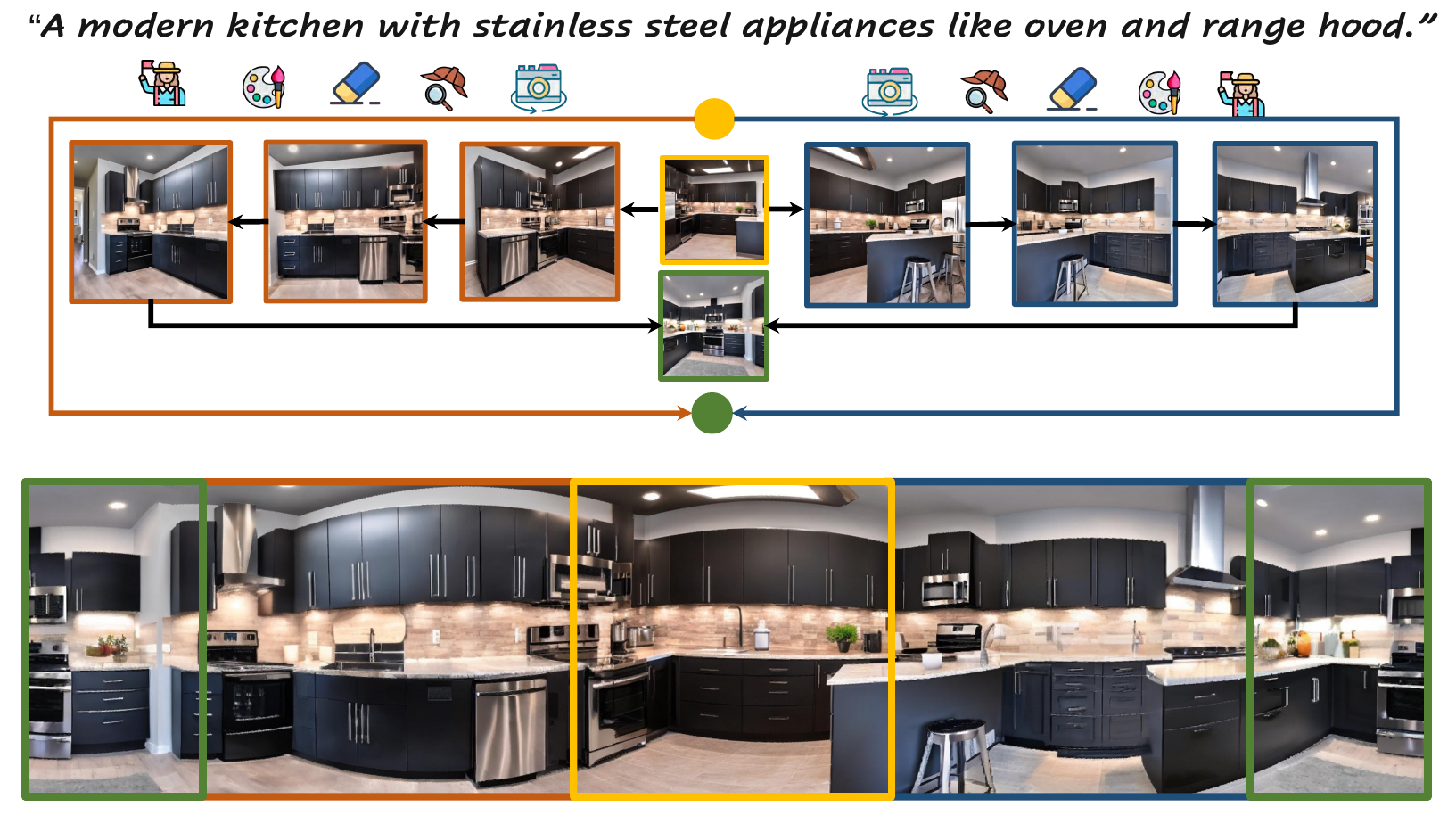}
  \vspace{-2mm}
  \caption{Detailed Illustration of 360° Panorama generation. }
  \label{fig:dil_360}
  \vspace{-2mm}
\end{figure*}

\noindent \textbf{360° Panorama Generation} is illustrated in Fig.~\ref{fig:dil_360}. Similarly, the Bidirectional Generation with Symmetric Guidance strategy is employed. Generation starts from the initial view centering (pitch, yaw) = (0°,0°), then undergoes symmetric rotation in two directions around the yaw axis. Finally, the two generation paths converge at the merging view with (pitch, yaw) = (0°,180°). Inpainting is used to merge the two generation paths to ensure loop closure.

\begin{figure*}[h]
  \centering
  \includegraphics[width=0.99\linewidth]{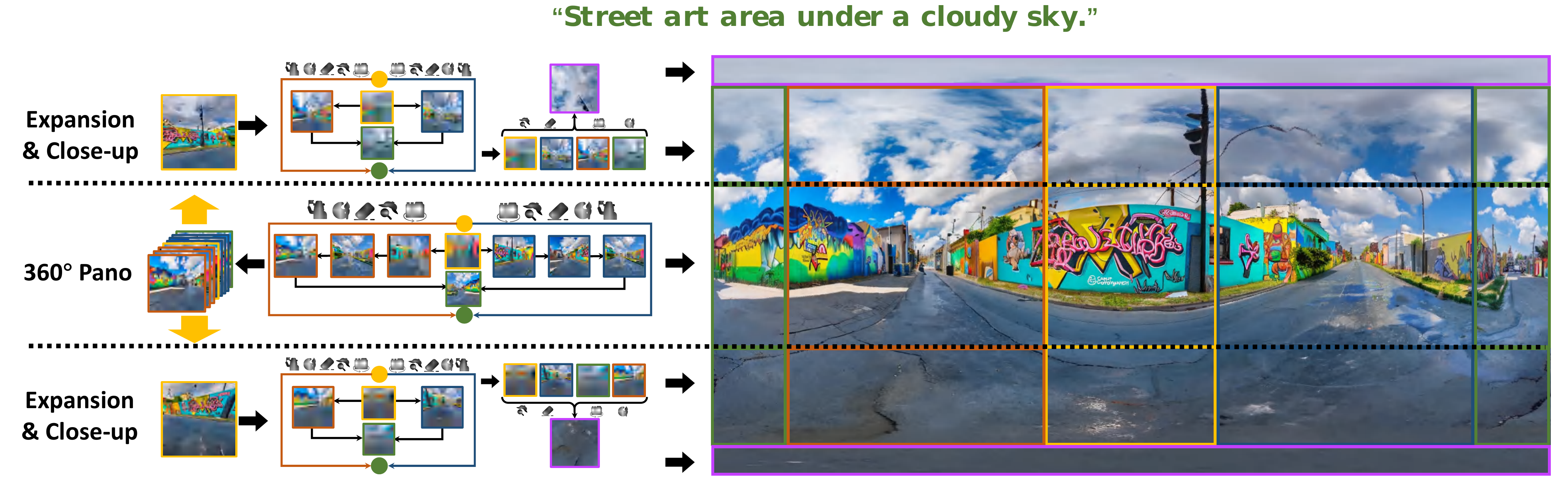}
  \vspace{-2mm}
  \caption{Detailed Illustration of Full Spherical Panorama generation. }
  \label{fig:dil_full}
  \vspace{-2mm}
\end{figure*}

\noindent \textbf{Full Spherical Generation} is illustrated in Fig.~\ref{fig:dil_full}. We firstly generate a 360° panorama, then expand in the upward and downward directions. Finally, we use inpainting at the upper and lower poles to close up the entire spherical panorama. Specifically, with the generated 360° panorama, we firstly warp to (pitch, yaw) = ($\phi$, 0°) and (pitch, yaw) = ($-\phi$, 0°) and inpaint the unknown areas to generate the initial views for the upward and downward expansions. Then, we apply PanoFree to expand the panorama's range in the pitch direction separately. Finally, we warp to (pitch, yaw) = (90°, 0°) and (pitch, yaw) = (-90°, 0°) and inpaint the unknown areas to close up the entire spherical panorama.

\subsection{Accumulated Errors in Vanilla Sequential Generation}
In this part, we provide detailed illustrations of the major accumulated errors that occur in the vanilla sequential generation process, and the risky erasing operations based on distance, edge color and smoothness.

\begin{figure*}[h]
  \centering
  \includegraphics[width=0.99\linewidth]{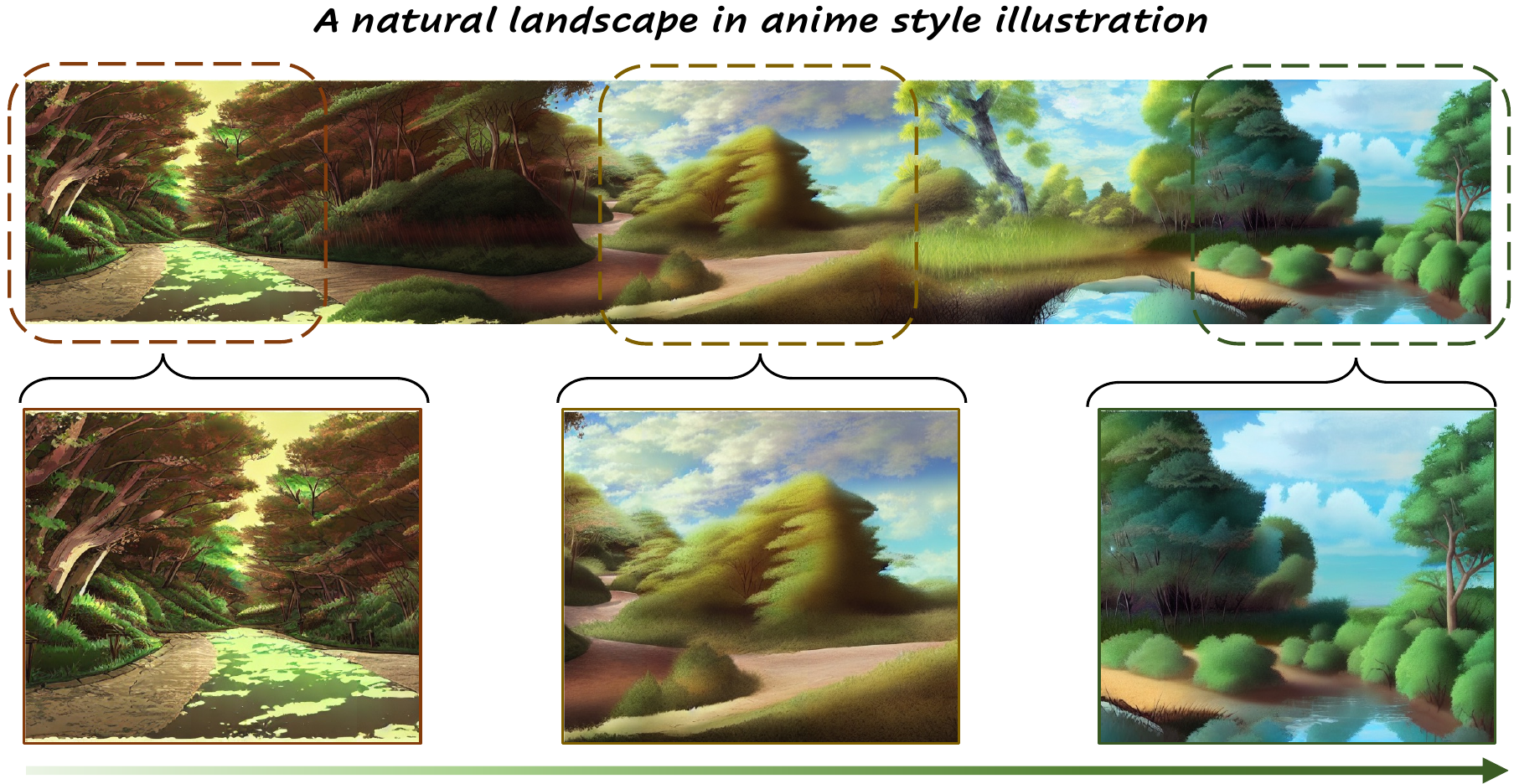}
  \vspace{-2mm}
  \caption{Illustration of accumulated inconsistency. }
  \label{fig:dil_inconsist}
  \vspace{-2mm}
\end{figure*}

\noindent \textbf{Accumulated Inconsistency} is illustrated in Fig.~\ref{fig:dil_inconsist}. 
As the vanilla sequential generation process solely condition the current view on the previous view, slight style and content shifts that occur during every warping and inpainting step may accumulate. This accumulation can lead to significant differences between distant regions, thereby damaging the global consistency of the generated panorama.

\begin{figure*}[h]
  \centering
  \includegraphics[width=0.99\linewidth]{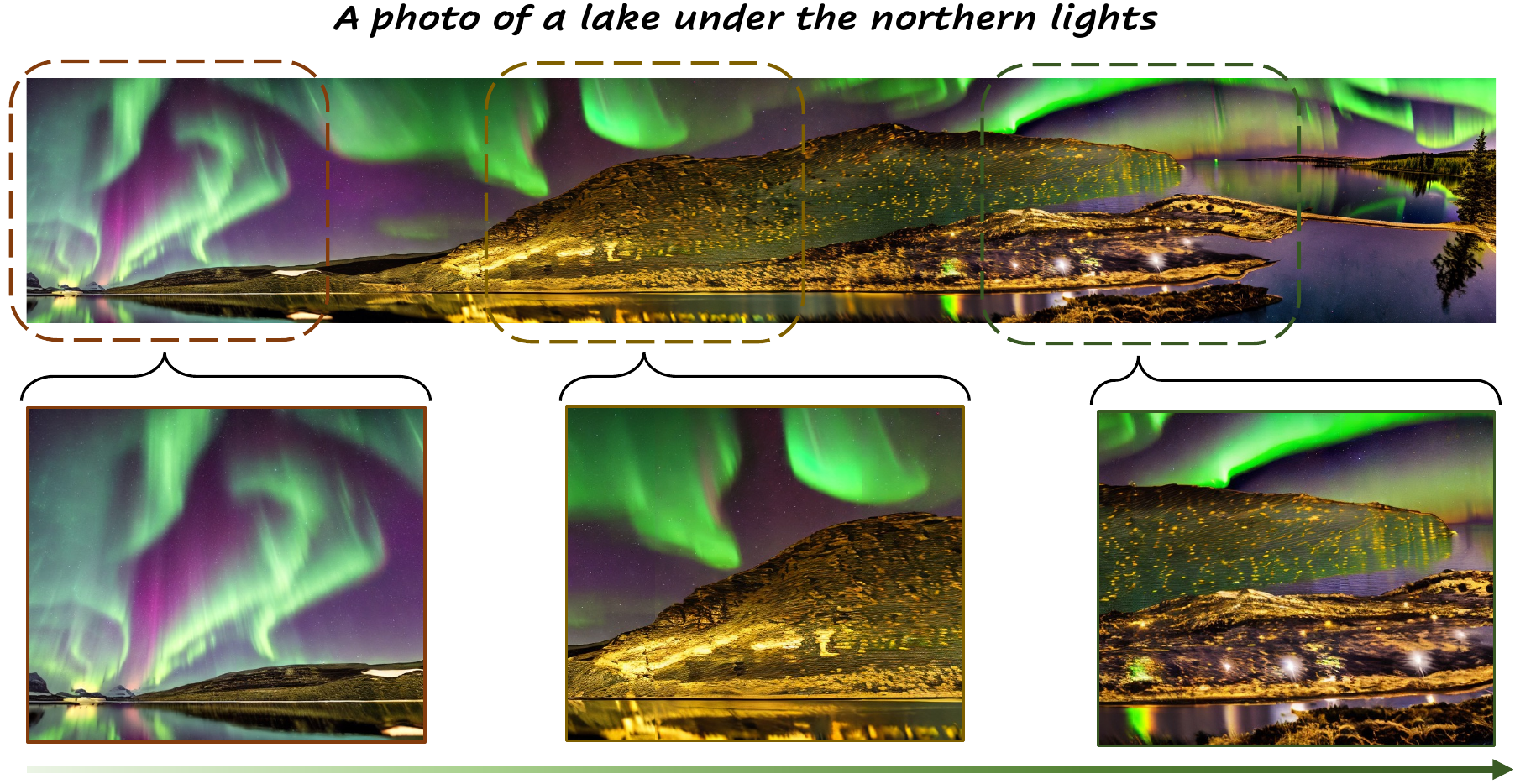}
  \vspace{-2mm}
  \caption{Illustration of accumulated artifacts. }
  \label{fig:dil_artifact}
  \vspace{-2mm}
\end{figure*}

\noindent \textbf{Accumulated Artifacts} are illustrated in Fig.~\ref{fig:dil_artifact}. In the vanilla sequential generation process, normal content in the current view may become artifacts in the following view after several warping and inpainting steps. Moreover, these artifacts can propagate with sequential generation, leading to the generation of more severe artifacts in subsequent views. Those contents that lead to artifacts is called as artifact-inducing contents.

\begin{figure}[h]
  \centering
  \begin{subfigure}{0.49\linewidth}
    \includegraphics[width=0.95\linewidth]{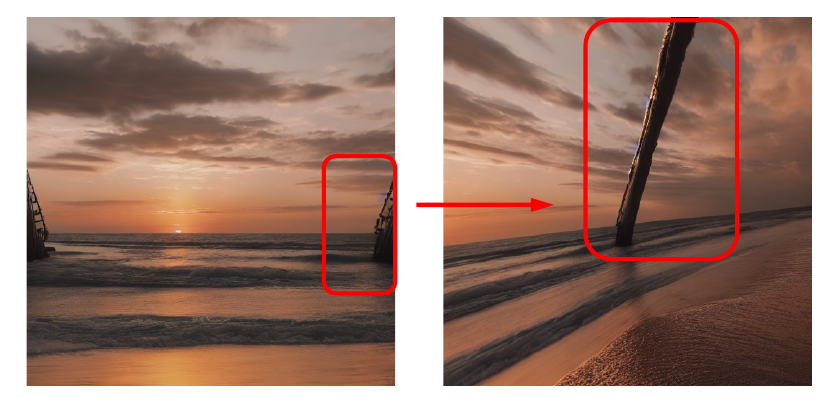}
    \caption{Truncated Objects}
    \label{fig:truncated_objects}
  \end{subfigure}
  \begin{subfigure}{0.49\linewidth}
    \includegraphics[width=0.93\linewidth]{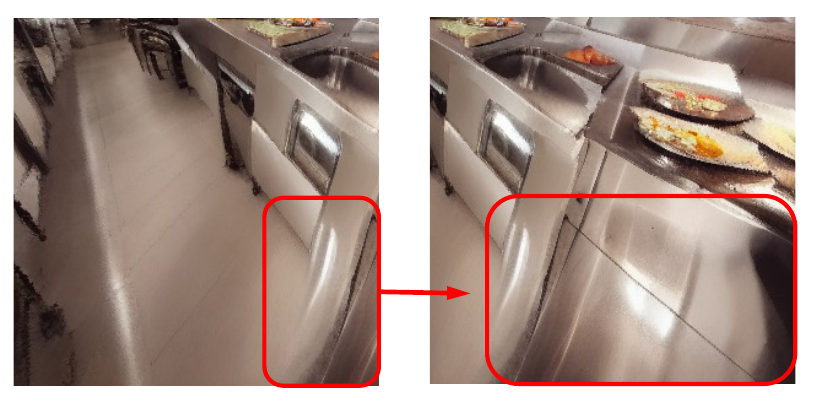}
    \caption{Distorted Areas}
    \label{fig:distorted_areas}
  \end{subfigure}
  \\
  \begin{subfigure}{0.48\linewidth}
    \includegraphics[width=0.93\linewidth]{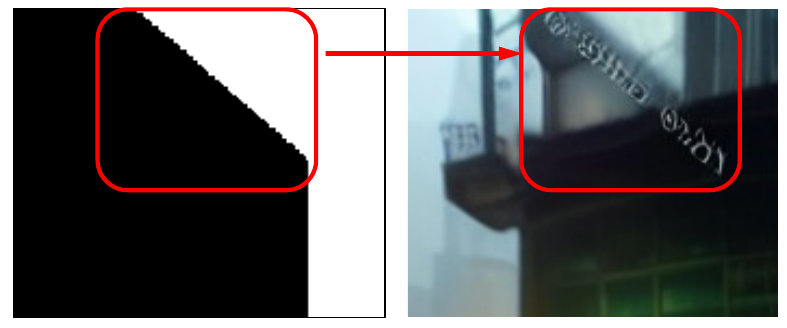}
    \caption{Jagged Edges}
    \label{fig:jagged_edges}
  \end{subfigure}
  \begin{subfigure}{0.48\linewidth}
    \includegraphics[width=0.93\linewidth]{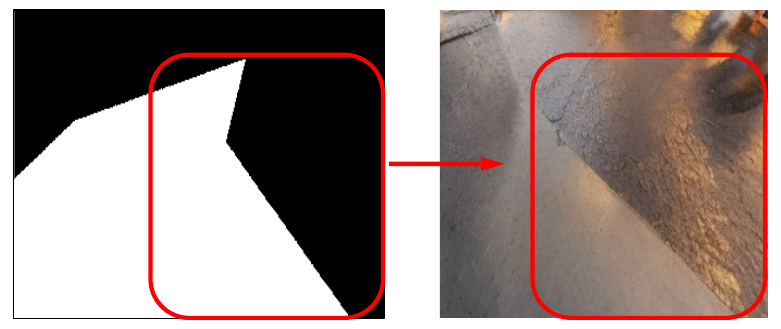}
    \caption{Sharp Edges}
    \label{fig:sharp_edges}
  \end{subfigure}
  \\
  \begin{subfigure}{0.97\linewidth}
    \includegraphics[width=0.99\linewidth]{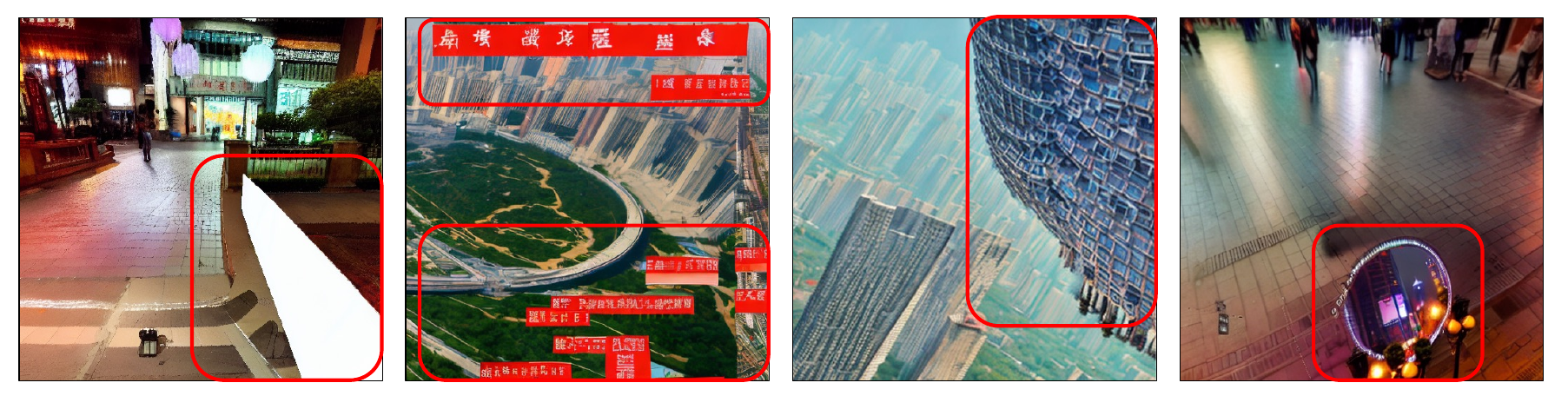}
  \caption{Salient areas with abrupt colors or unevenness.}
  \label{fig:salient_areas}
  \end{subfigure}
  
  \caption{Illustration of artifact-inducing contents.}
  \label{fig:artifact_inducing}
\end{figure}

\noindent \textbf{Artifact-inducing Contents} are illustrated in Fig.~\ref{fig:artifact_inducing}. The major artifact-inducing contents and the artifacts they cause are summarized as follows:
\begin{itemize}
  \item {\it Truncated Objects} are illustrated in  Fig.~\ref{fig:truncated_objects}. We observed that pretrained T2I models often generate partial objects truncated by the edges of each view. Those objects can be extended into unreasonable contents in the following views during the warping and inpainting process.

  \item {\it Distorted Areas} are illustrated in  Fig.~\ref{fig:distorted_areas}. Regions heavily distorted during warping may appear blurred or exhibit strange shapes. After inpainting, these distorted regions may be extended, resulting in a large distorted area on the panorama. Those distorted areas often occur near the edges of each view.
  
  \item {\it Jagged Edges} are illustrated in  Fig.~\ref{fig:jagged_edges}. Jagged edges on the inpainting masks often lead to jagged artifacts or letter-like artifacts.
   
  \item {\it Sharp Edges} are illustrated in  Fig.~\ref{fig:sharp_edges}. Sharp edges on the inpainting masks often result in inconsistent connections between inpainted regions and other regions of the image, causing noticeable boundaries or color discrepancies.
  
  \item {\it Salient Areas} with abrupt colors or unevenness are illustrated in Fig.~\ref{fig:salient_areas}. These salient areas may appear reasonable under the current view, but after warping and inpainting, they can easily accumulate into noticeable artifacts. On the other hand, abrupt colors or unevenness are already characteristic features of artifacts.
\end{itemize}

\begin{figure}[h]
  \centering
  \includegraphics[width=0.99\linewidth]{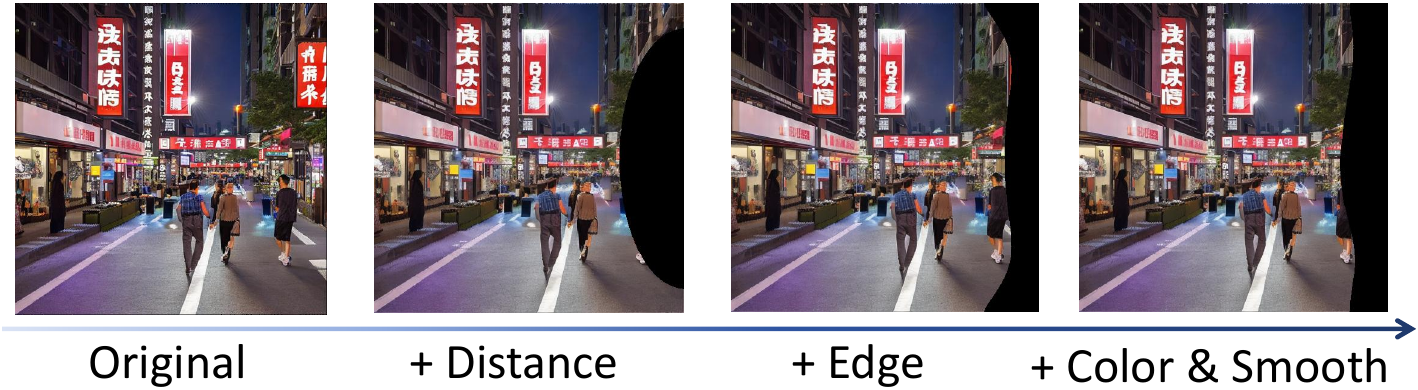}
  \caption{Illustration of Risky Area Erasing.}
  \label{fig:ill_rae}
\end{figure}

\noindent \textbf{Risky Area Erasing} are briefly illustrated in Fig.~\ref{fig:ill_rae}, in which distance-based erasing, edge-based erasing, and color \& smoothness-based erasing are progressively applied and black regions represent the areas that have been erased.

\begin{figure*}[ht]
  \centering
  \includegraphics[width=0.99\linewidth]{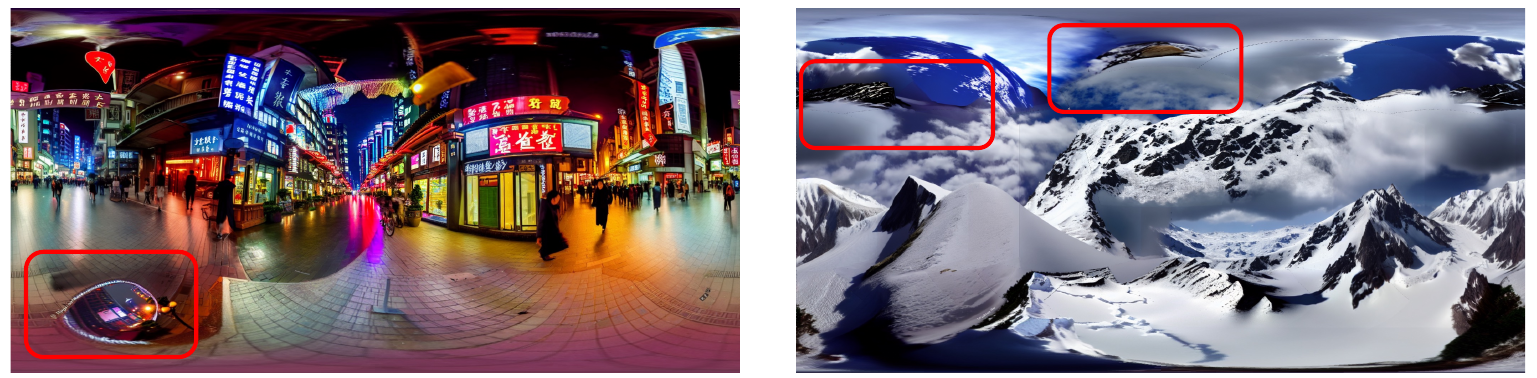}
  \vspace{-2mm}
  \caption{Examples of Hallucinations.}
  \label{fig:dil_hallucinations}
  \vspace{-2mm}
\end{figure*}

\noindent \textbf{Hallucinations} are illustrated in Fig.~\ref{fig:dil_hallucinations}. Hallucinations often occur when generating full spherical panoramas. When we use the same text description to generate all views on the spherical panorama, it may result in conflicts between the generated content's placement and the scene structure prior. For instance, another city on the ground or a mountain peak floating in the sky could be generated.

\section{Ablation Study}

We furthr conducted ablation studies to evaluate the impact of different components of PanoFree and guidance scale respectively. 

\begin{figure}[h]
  \centering
  \includegraphics[width=0.99\linewidth]{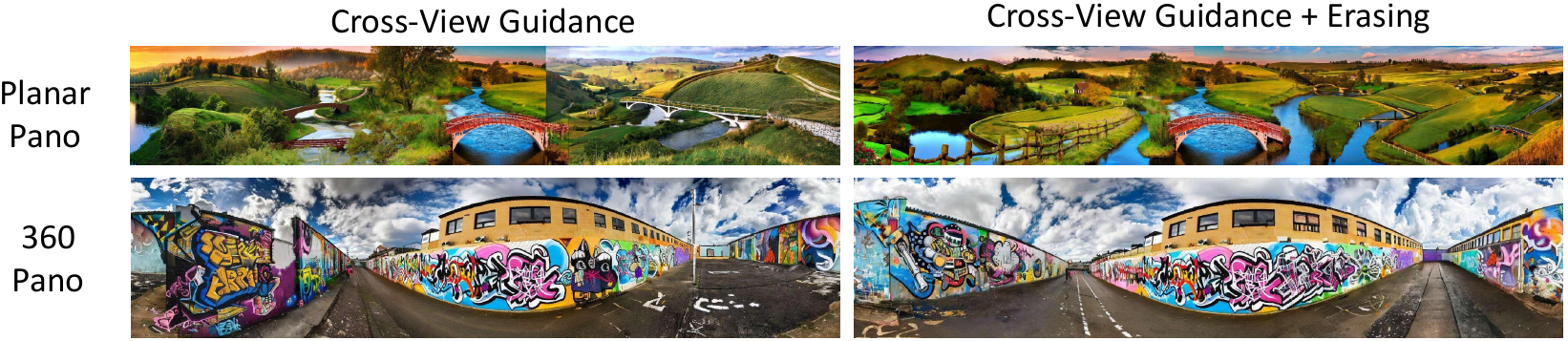}
  \caption{Qualitative ablation of each component in PanoFree.}
  \label{fig:ab_imgs}
  \vspace{-1mm}
\end{figure}

 
 \noindent\textbf{Qualitative Ablation.} Fig.\ref{fig:ab_imgs} showcases representative examples for qualitative ablation.  Qualitative results support the prior observations in quantitative ablation, demonstrating progressive improvement in image coherence and quality with each integrated component.

\subsection{Ablation of Guidance Scale}
\begin{figure}[h]
  \centering
  \vspace{-1.5mm}
  \includegraphics[width=0.99\linewidth]{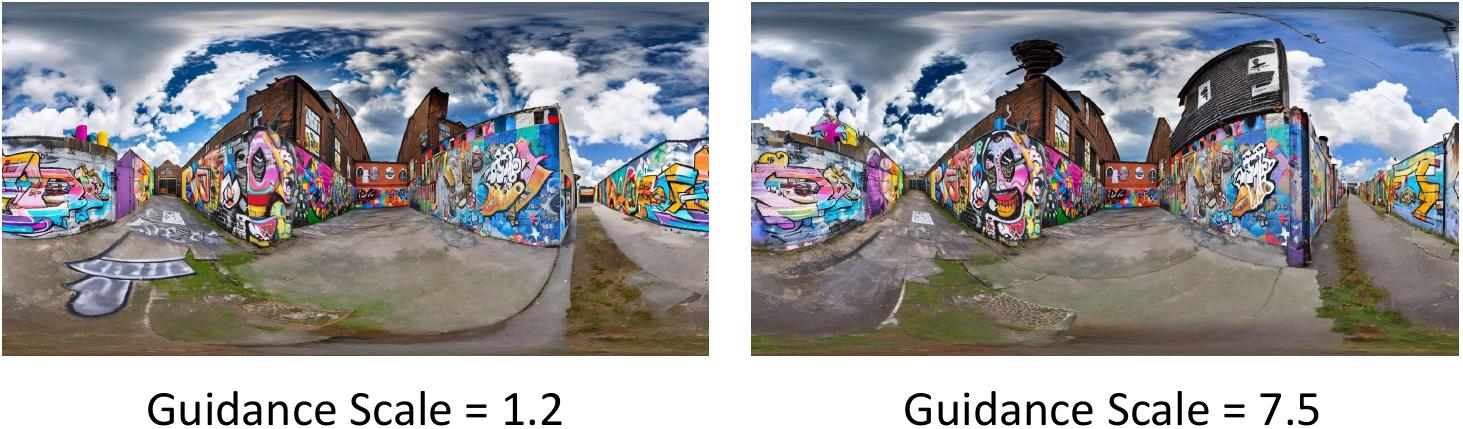}
    \vspace{-2mm}
    \caption{Qualitative ablation of guidance scales.}
    \label{fig:ab_gfs}
  \vspace{-3mm}
\end{figure}

As the guidance scale primarily affects full panorama generation, we perform a qualitative ablation in Fig.~\ref{fig:ab_gfs}, showing that large guidance scales during expansion and close-up stages can cause artifacts, like the unusual structures in the upper part of the right image.

\section{Additional Experiment Results and Comparisons}
In this section, we present additional experimental results, including the Planar, 360°, and Full Spherical Panoramas generated by PanoFree. We categorize these results into indoor, street and city, as well as natural scenes, to comprehensively showcase PanoFree's generation capability. Furthermore, we provide further comparisons with baseline methods. 
\subsection{Planar Panorama Generation}
\begin{figure*}[h]
  \centering
  \includegraphics[width=0.99\linewidth]{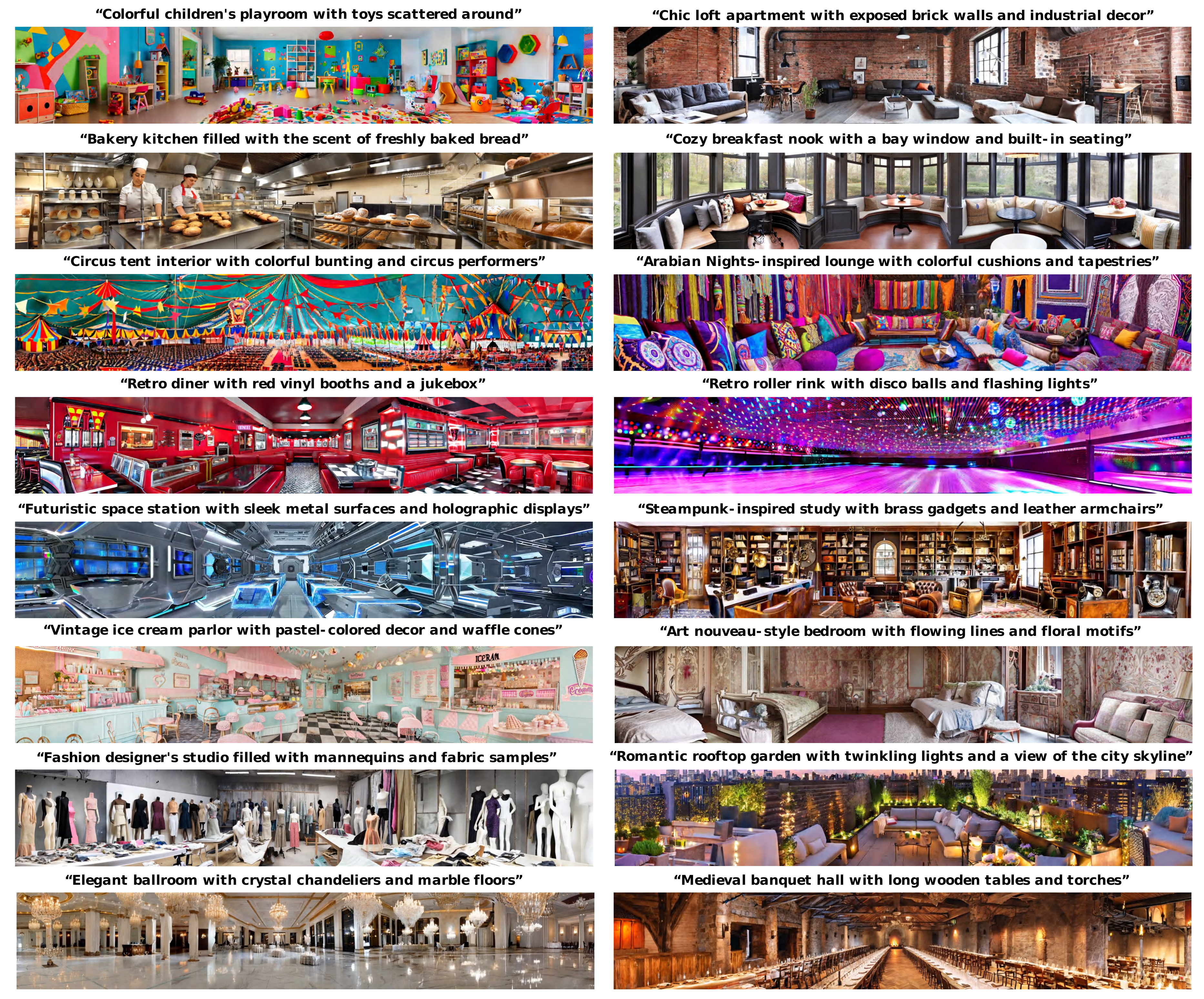}
  \vspace{-2mm}
  \caption{Indoor scene planar panoramas generated by PanoFree.}
  \label{fig:pf_planar_indoor}
  \vspace{-2mm}
\end{figure*}

\begin{figure*}[h]
  \centering
  \includegraphics[width=0.99\linewidth]{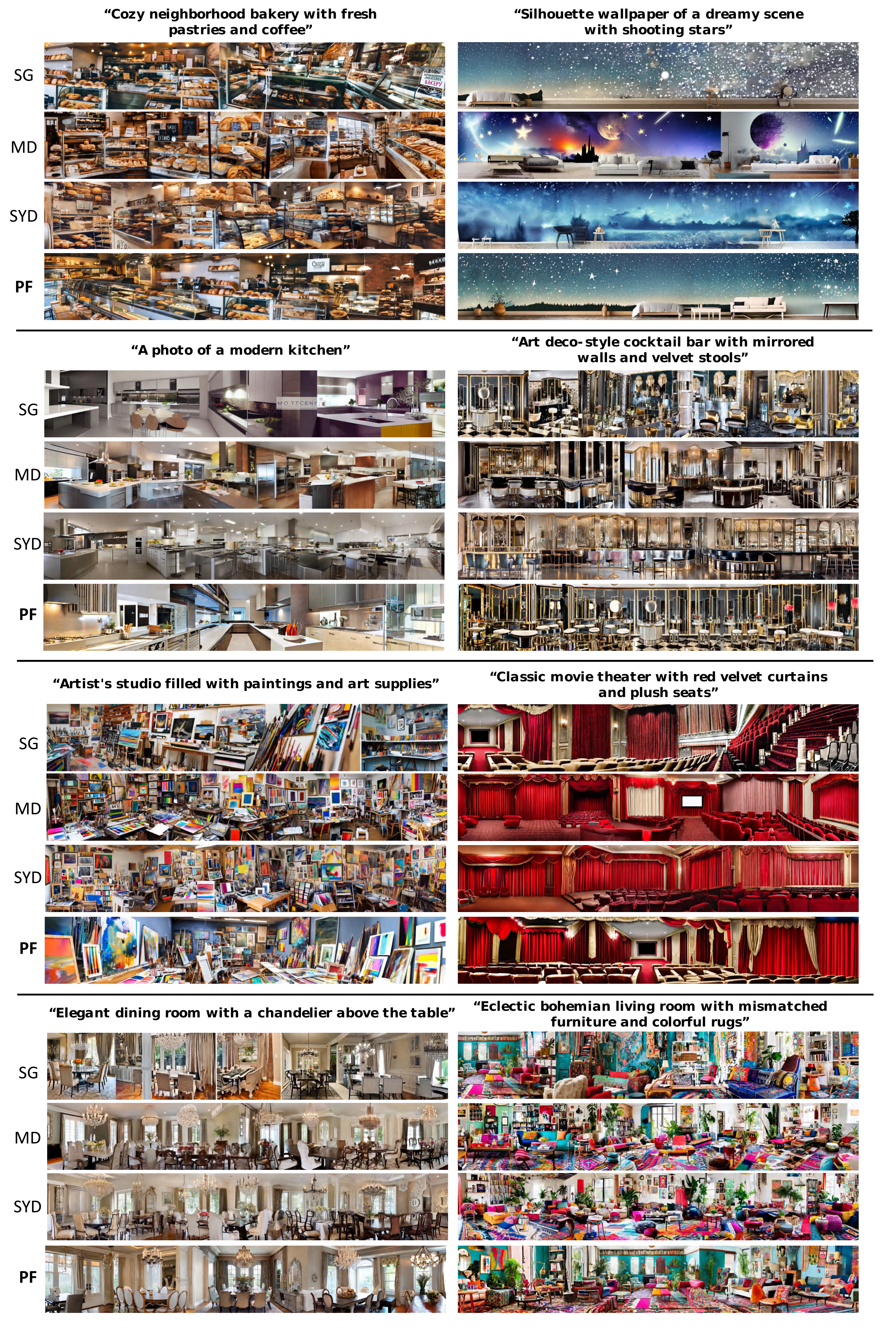}
  \vspace{-2mm}
  \caption{Comparison: Indoor scene planar panoramas. PF is our PanoFree.}
  \label{fig:comp_planar_indoor}
  \vspace{-2mm}
\end{figure*}

\noindent \textbf{Indoor Scene} panoramas generated with PanoFree are shown in  Fig.~\ref{fig:pf_planar_indoor}. 
We also provide additional comparisons with vanilla sequential generation (SG), MultiDiffusion (MD)~\cite{bar2023multidiffusion}, and SyncDiffusion (SYD)~\cite{lee2024syncdiffusion} in Fig.~\ref{fig:comp_planar_indoor}. 

\begin{figure*}[h]
  \centering
  \includegraphics[width=0.99\linewidth]{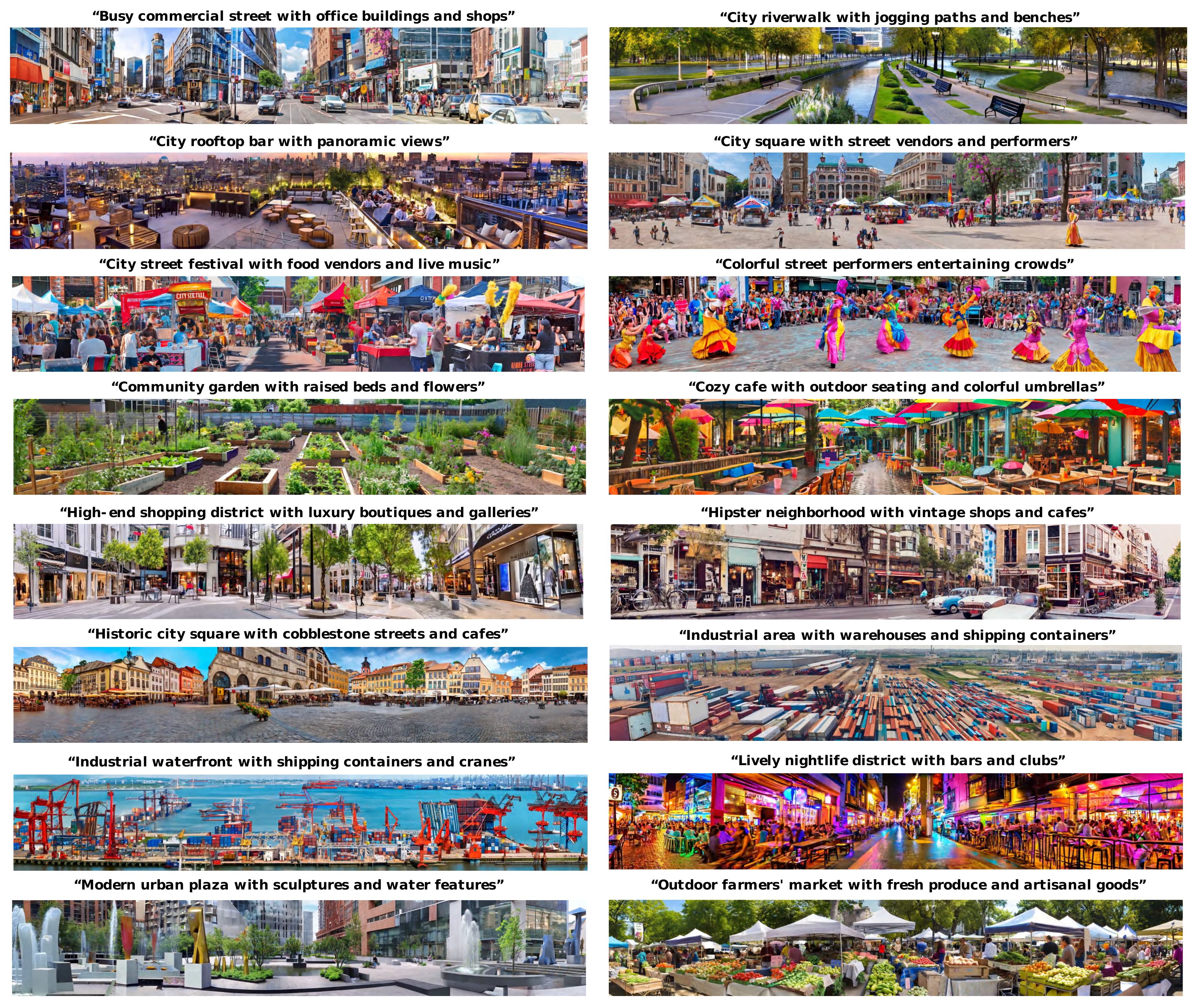}
  \vspace{-2mm}
  \caption{City and street scene planar panoramas generated by PanoFree.}
  \label{fig:pf_planar_city}
  \vspace{-2mm}
\end{figure*}

\begin{figure*}[h]
  \centering
  \includegraphics[width=0.99\linewidth]{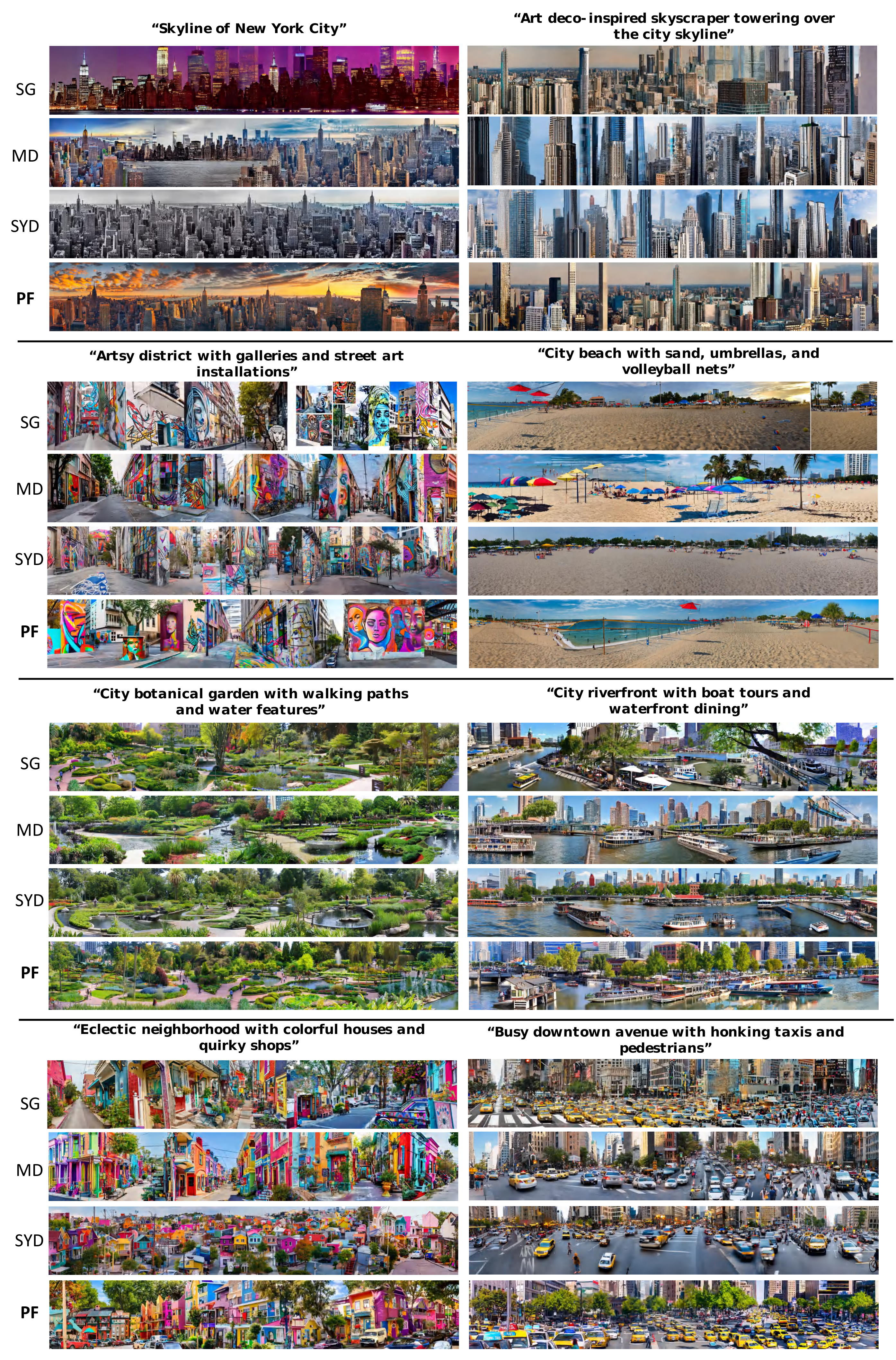}
  \vspace{-2mm}
  \caption{Comparison: City and street scene planar panoramas. PF is our PanoFree.}
  \label{fig:comp_planar_city}
  \vspace{-2mm}
\end{figure*}

\noindent \textbf{City and Street Scene} panoramas generated with PanoFree are shown in  Fig.~\ref{fig:pf_planar_city}. 
We also provide additional comparisons with vanilla sequential generation (SG), MultiDiffusion (MD)~\cite{bar2023multidiffusion}, and SyncDiffusion (SYD)~\cite{lee2024syncdiffusion} in Fig.~\ref{fig:comp_planar_city}. 

\begin{figure*}[h]
  \centering
  \includegraphics[width=0.99\linewidth]{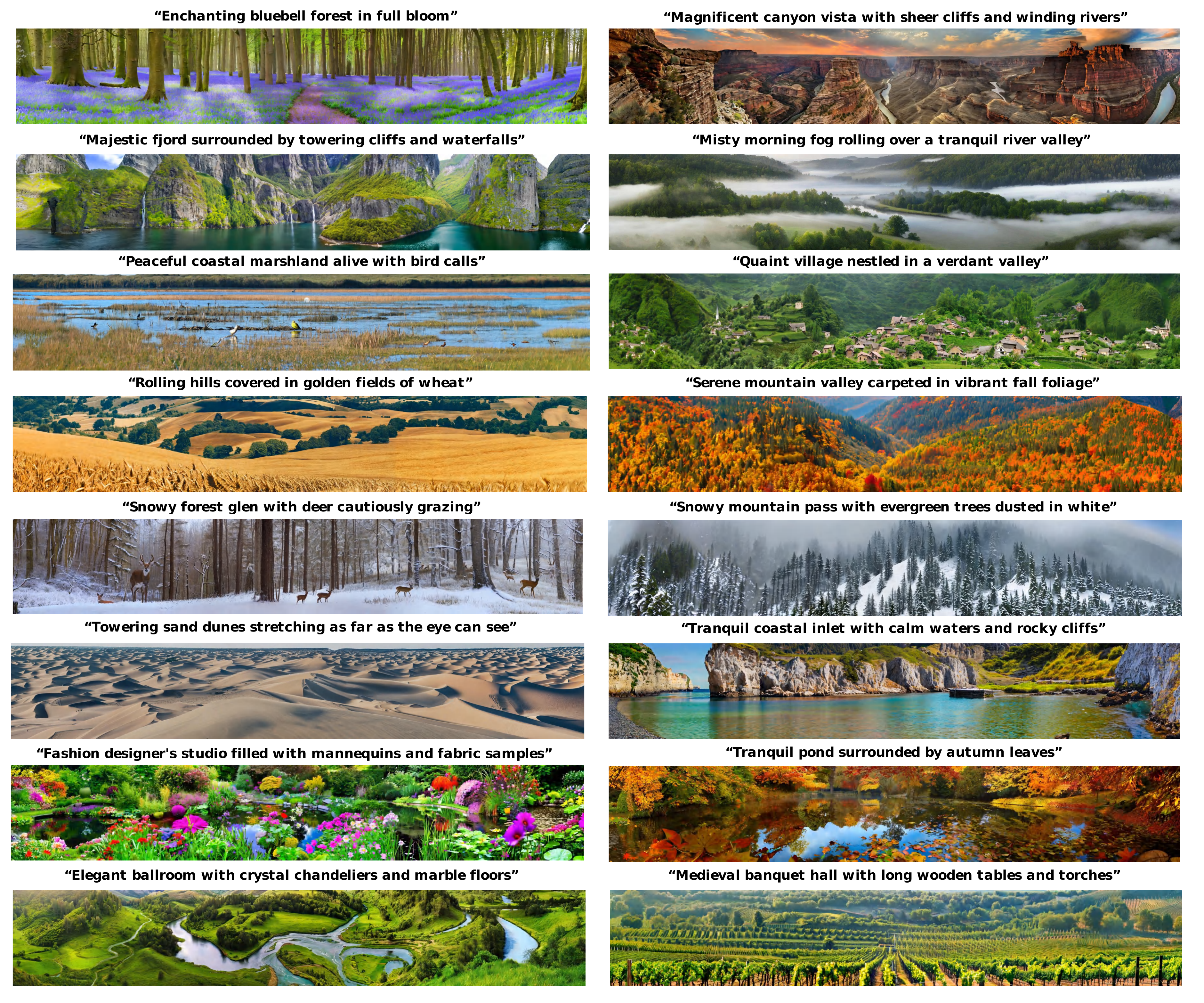}
  \vspace{-2mm}
  \caption{Natural scene planar panoramas generated by PanoFree.}
  \label{fig:pf_planar_natural}
  \vspace{-2mm}
\end{figure*}

\begin{figure*}[h]
  \centering
  \includegraphics[width=0.99\linewidth]{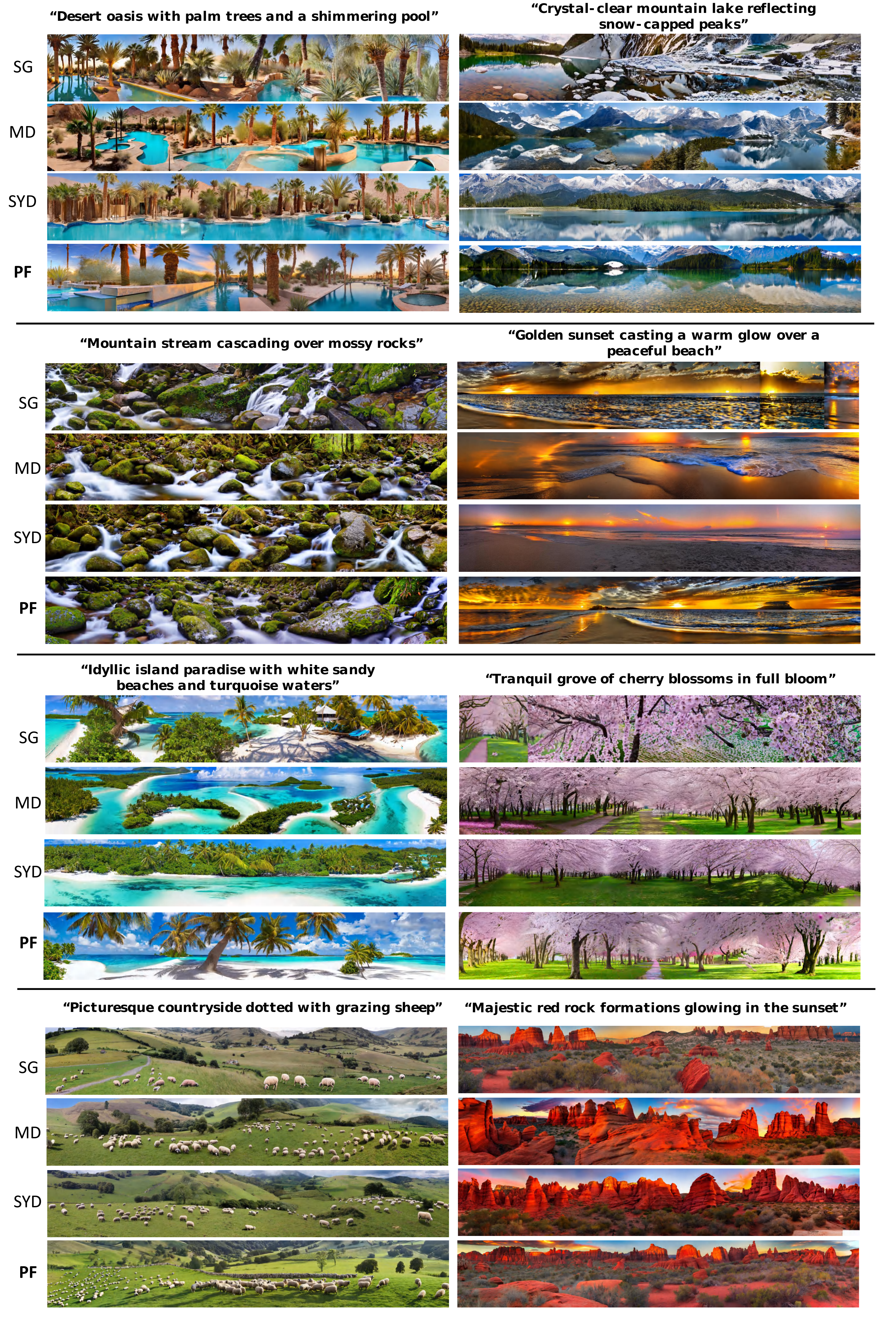}
  \vspace{-2mm}
  \caption{Comparison: Natural scene planar panoramas. PF is our PanoFree.}
  \label{fig:comp_planar_natural}
  \vspace{-2mm}
\end{figure*}

\noindent \textbf{Natural Scene} panoramas generated with PanoFree are shown in  Fig.~\ref{fig:pf_planar_natural}. 
We also provide additional comparisons with vanilla sequential generation (SG), MultiDiffusion (MD)~\cite{bar2023multidiffusion}, and SyncDiffusion (SYD)~\cite{lee2024syncdiffusion} in Fig.~\ref{fig:comp_planar_natural}. 

We can observe that PanoFree is capable of generating  panoramas with various scenes, styles and contents. In terms of image quality and global consistency, PanoFree significantly outperforms vanilla sequential generation and MultiDiffusion and it's comparable to SyncDiffusion.

\subsection{360° Panorama Generation}

\begin{figure*}[h]
  \centering
  \includegraphics[width=0.99\linewidth]{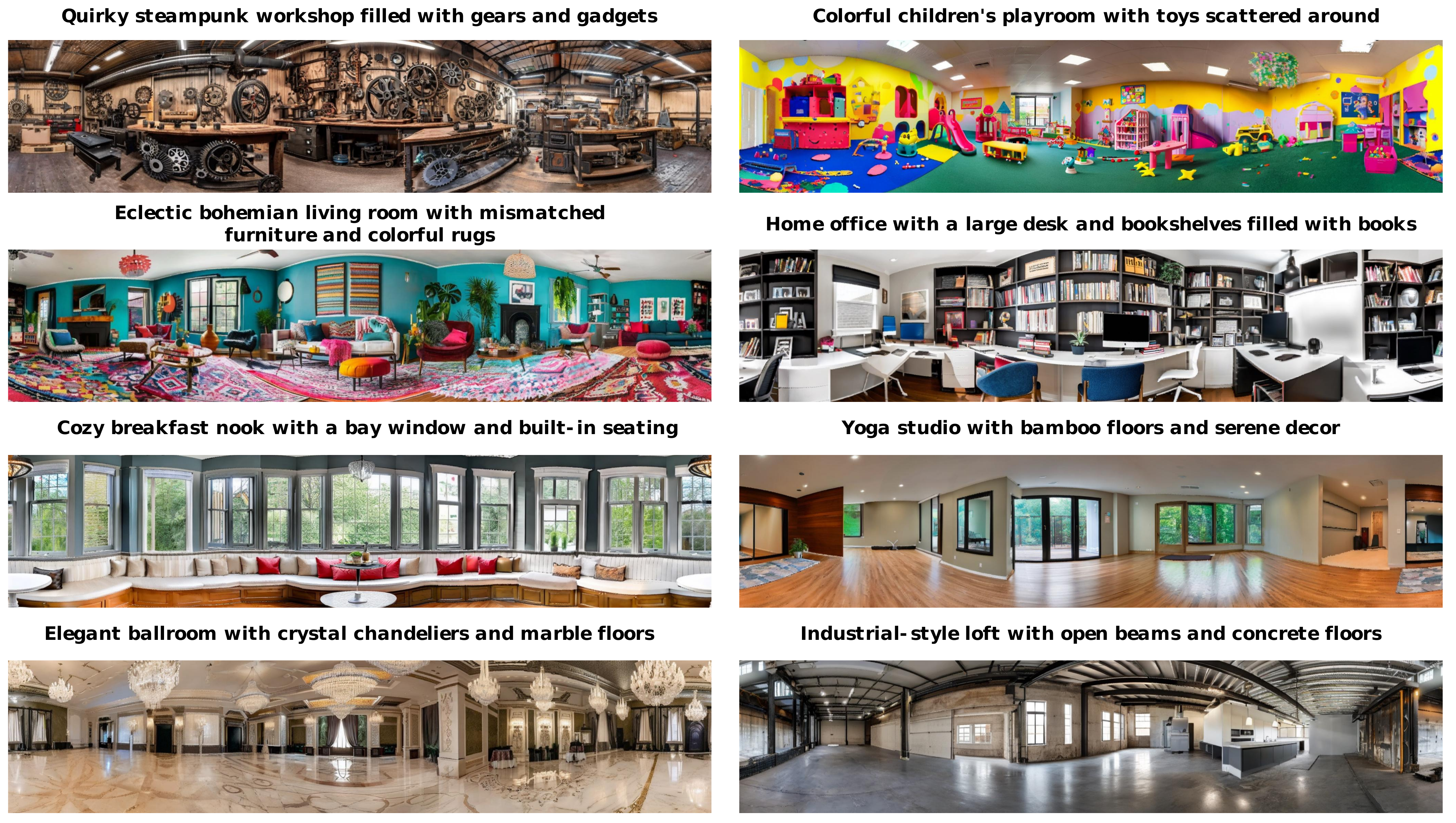}
  \vspace{-2mm}
  \caption{Indoor scene 360° panoramas generated by PanoFree.}
  \label{fig:pf_360_indoor}
  \vspace{-2mm}
\end{figure*}

\begin{figure*}[h]
  \centering
  \includegraphics[width=0.99\linewidth]{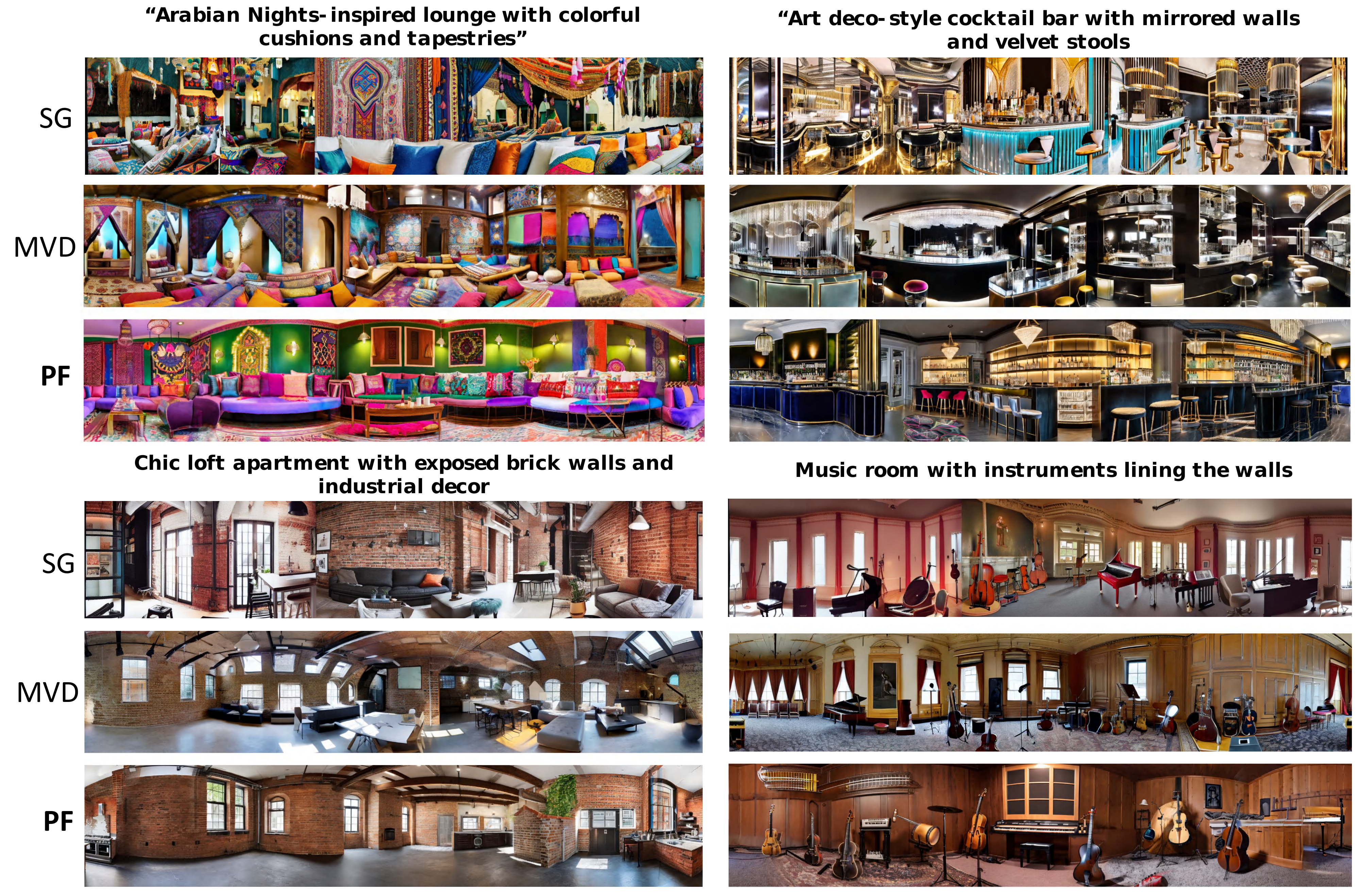}
  \vspace{-2mm}
  \caption{Comparison: Indoor scene 360° panoramas. PF is our PanoFree.}
  \label{fig:comp_360_indoor}
  \vspace{-2mm}
\end{figure*}

\noindent \textbf{Indoor Scene} panoramas generated with PanoFree are shown in  Fig.~\ref{fig:pf_360_indoor}. 
We also provide additional comparisons with vanilla sequential generation (SG) and MVDiffusion (MVD)~\cite{Tang2023MVDiffusionEH} in Fig.~\ref{fig:comp_360_indoor}. 

\begin{figure*}[h]
  \centering
  \includegraphics[width=0.99\linewidth]{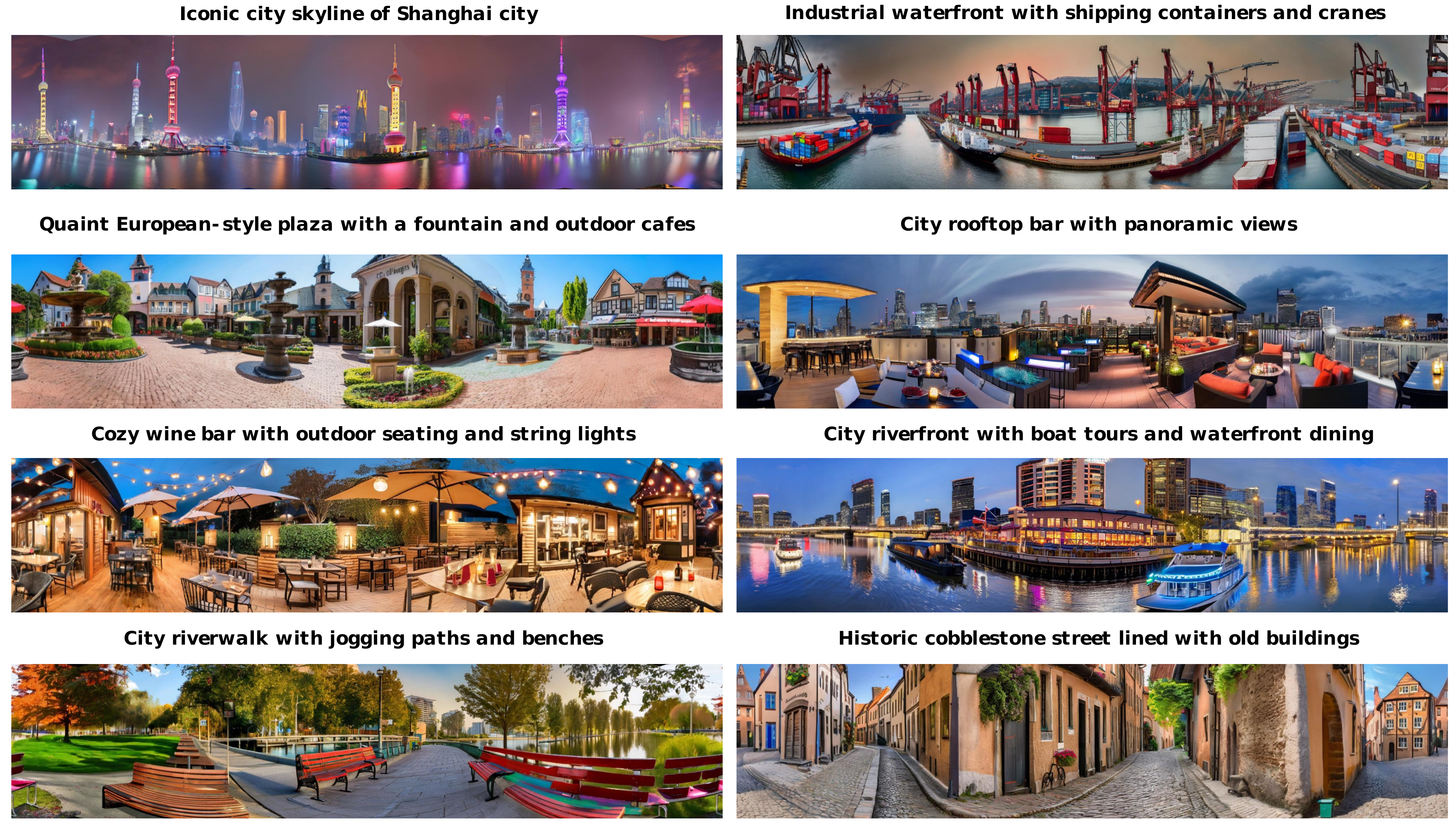}
  \vspace{-2mm}
  \caption{City and Street scene 360° panoramas generated by PanoFree.}
  \label{fig:pf_360_city}
  \vspace{-2mm}
\end{figure*}

\begin{figure*}[h]
  \centering
  \includegraphics[width=0.99\linewidth]{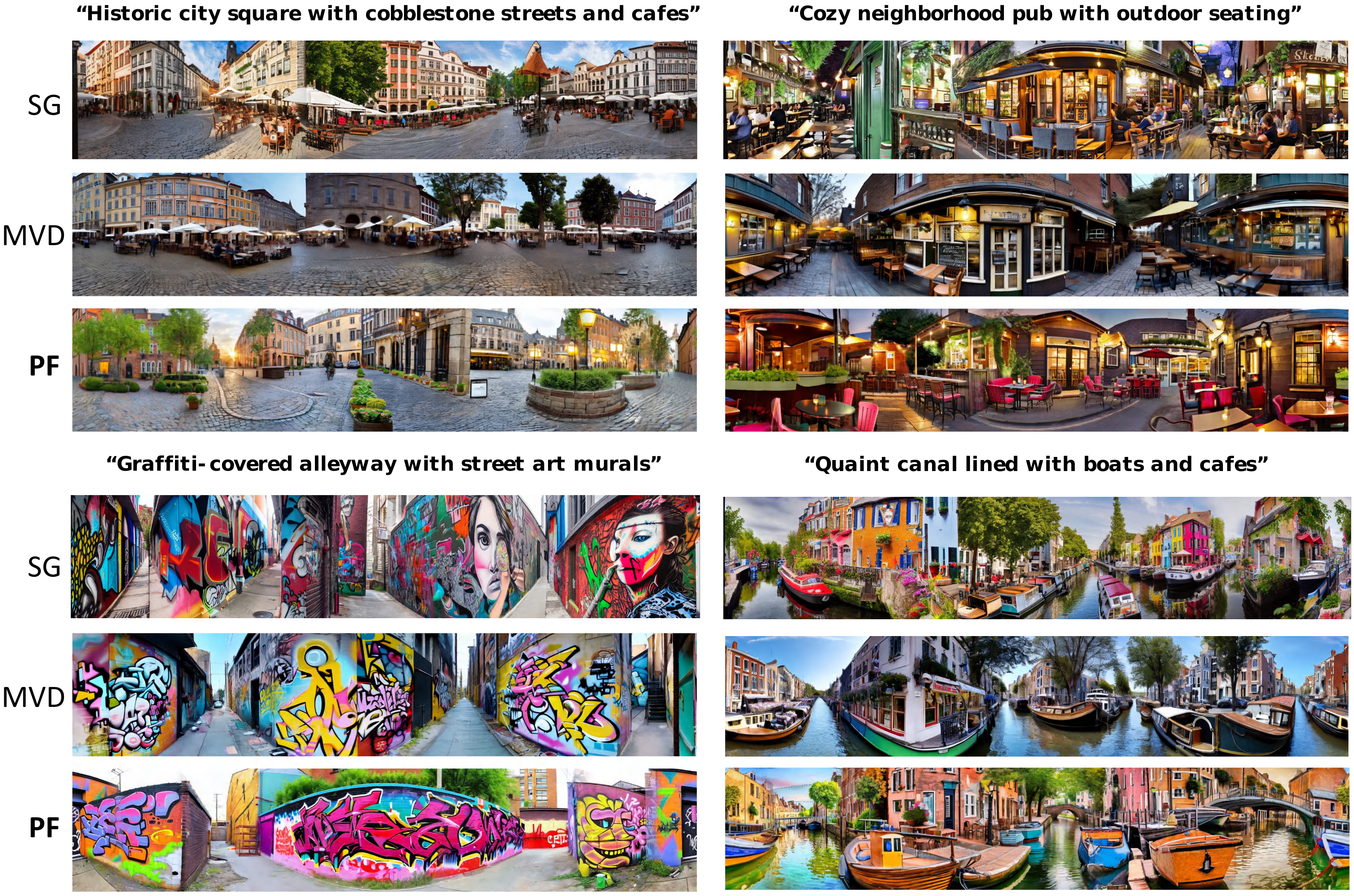}
  \vspace{-2mm}
  \caption{Comparison: City and Street scene 360° panoramas. PF is our PanoFree.}
  \label{fig:comp_360_city}
  \vspace{-2mm}
\end{figure*}

\noindent \textbf{City and Street Scene} panoramas generated with PanoFree are shown in Fig.~\ref{fig:pf_360_city}. We also provide additional comparisons with vanilla sequential generation (SG) and MVDiffusion (MVD)~\cite{Tang2023MVDiffusionEH} in Fig.~\ref{fig:comp_360_city}. 

\begin{figure*}[h]
  \centering
  \includegraphics[width=0.99\linewidth]{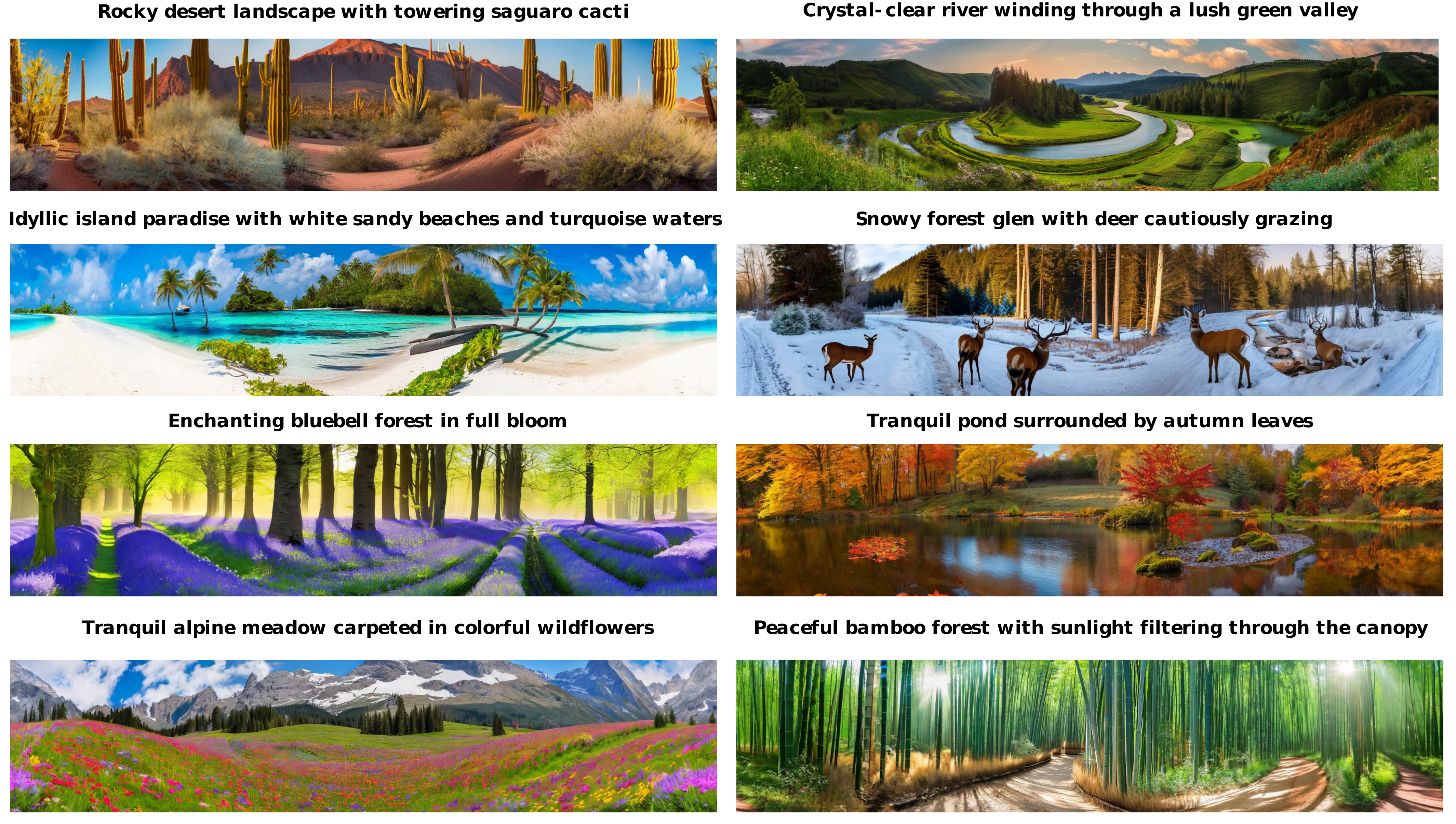}
  \vspace{-2mm}
  \caption{Natural scene 360° panoramas generated by PanoFree.}
  \label{fig:pf_360_natural}
  \vspace{-2mm}
\end{figure*}

\begin{figure*}[h]
  \centering
  \includegraphics[width=0.99\linewidth]{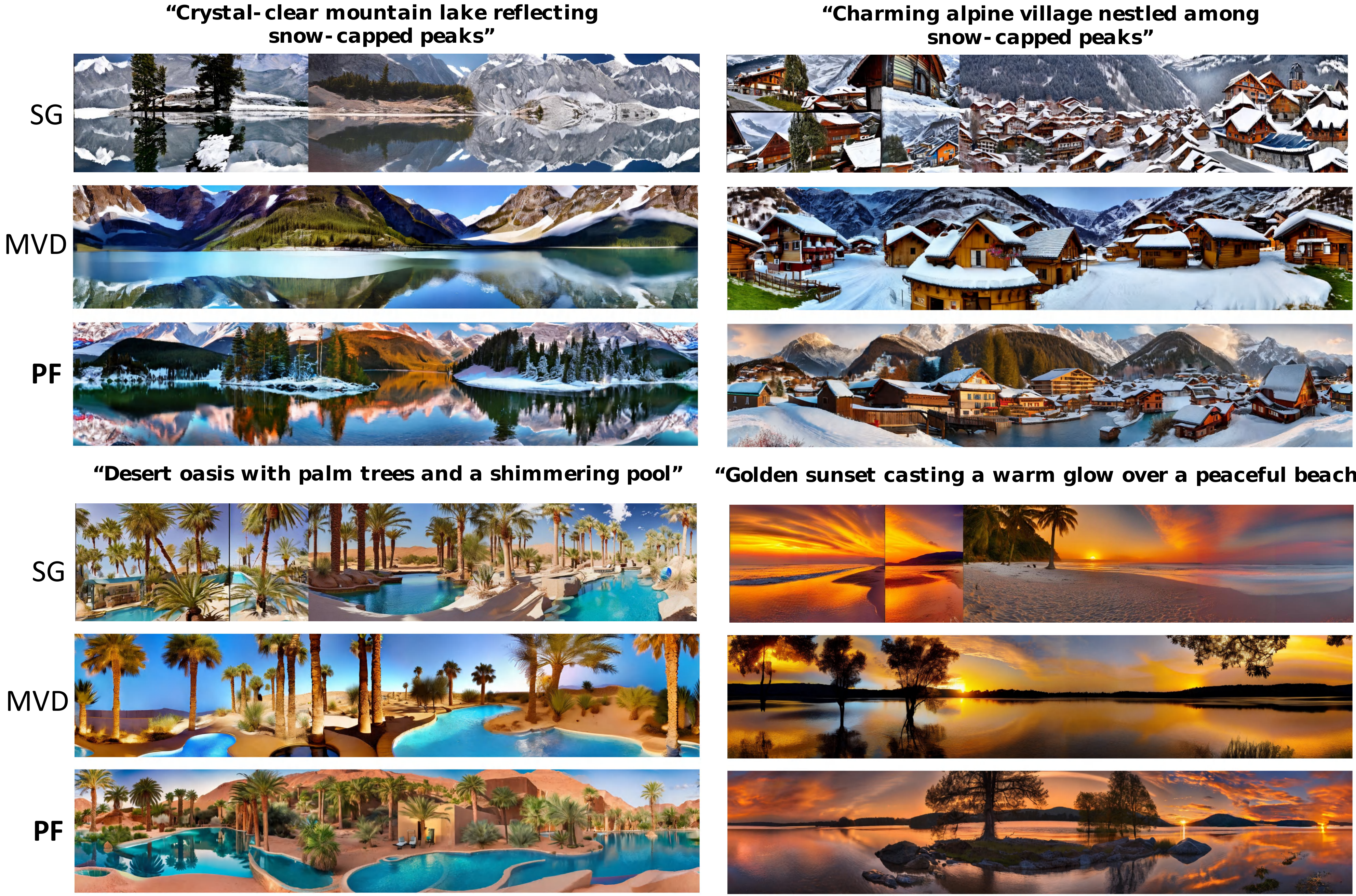}
  \vspace{-2mm}
  \caption{Comparison: Natural scene 360° panoramas. PF is our PanoFree.}
  \label{fig:comp_360_natural}
  \vspace{-2mm}
\end{figure*}

\noindent \textbf{Natural Scene} panoramas generated with PanoFree are shown in  Fig.~\ref{fig:pf_360_natural}. 
We also provide additional comparisons with vanilla sequential generation (SG) and MVDiffusion (MVD)~\cite{Tang2023MVDiffusionEH} in Fig.~\ref{fig:comp_360_natural}. 

\subsection{Full Spherical Generation}

\begin{figure*}[h]
  \centering
  \includegraphics[width=0.99\linewidth]{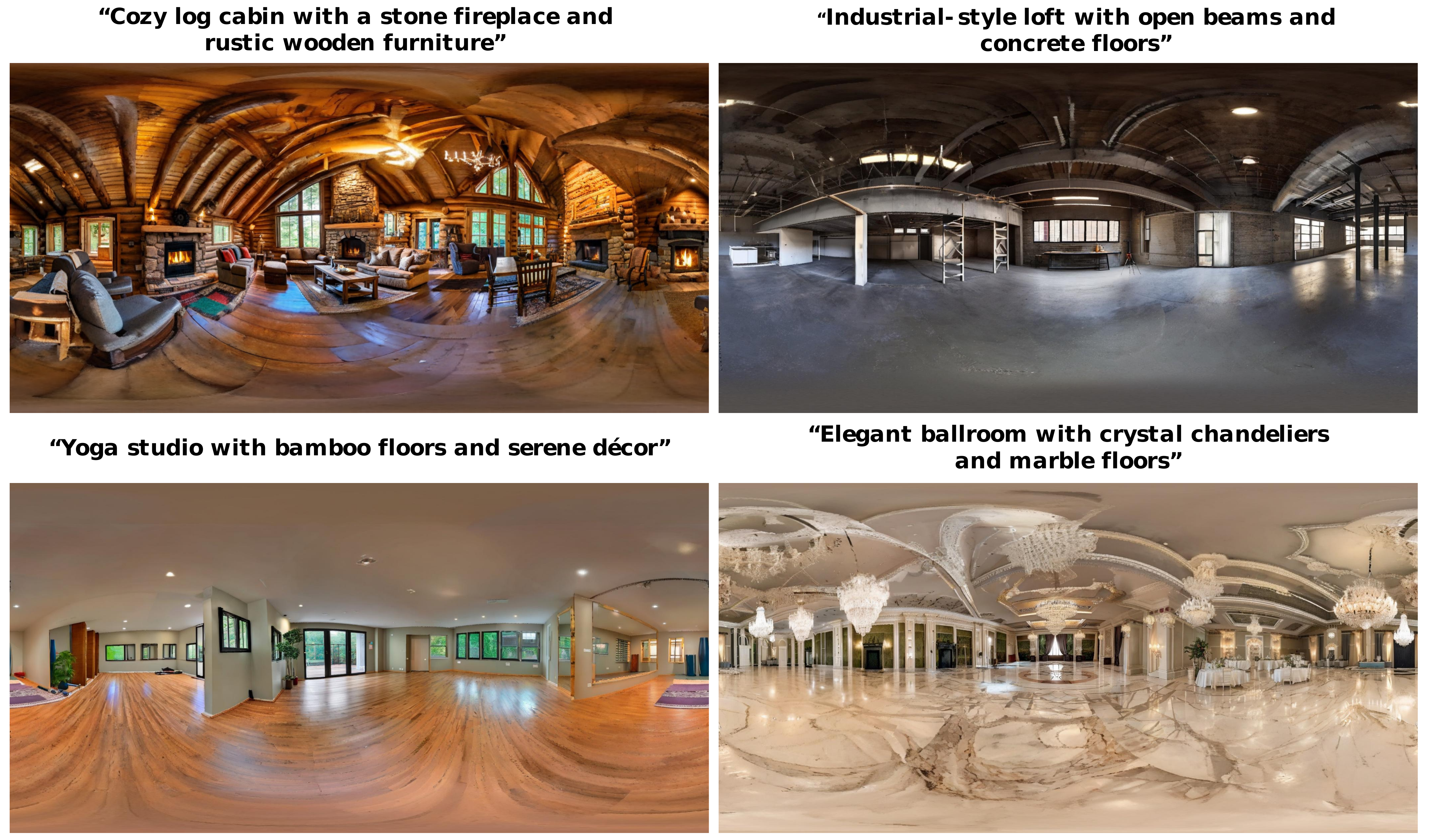}
  \vspace{-2mm}
  \caption{Indoor scene full spherical panoramas generated by PanoFree.}
  \label{fig:pf_full_indoor}
  \vspace{-2mm}
\end{figure*}

\noindent \textbf{Indoor Scene} panoramas generated with PanoFree are shown in Fig.~\ref{fig:pf_full_indoor}.

\begin{figure*}[h]
  \centering
  \includegraphics[width=0.99\linewidth]{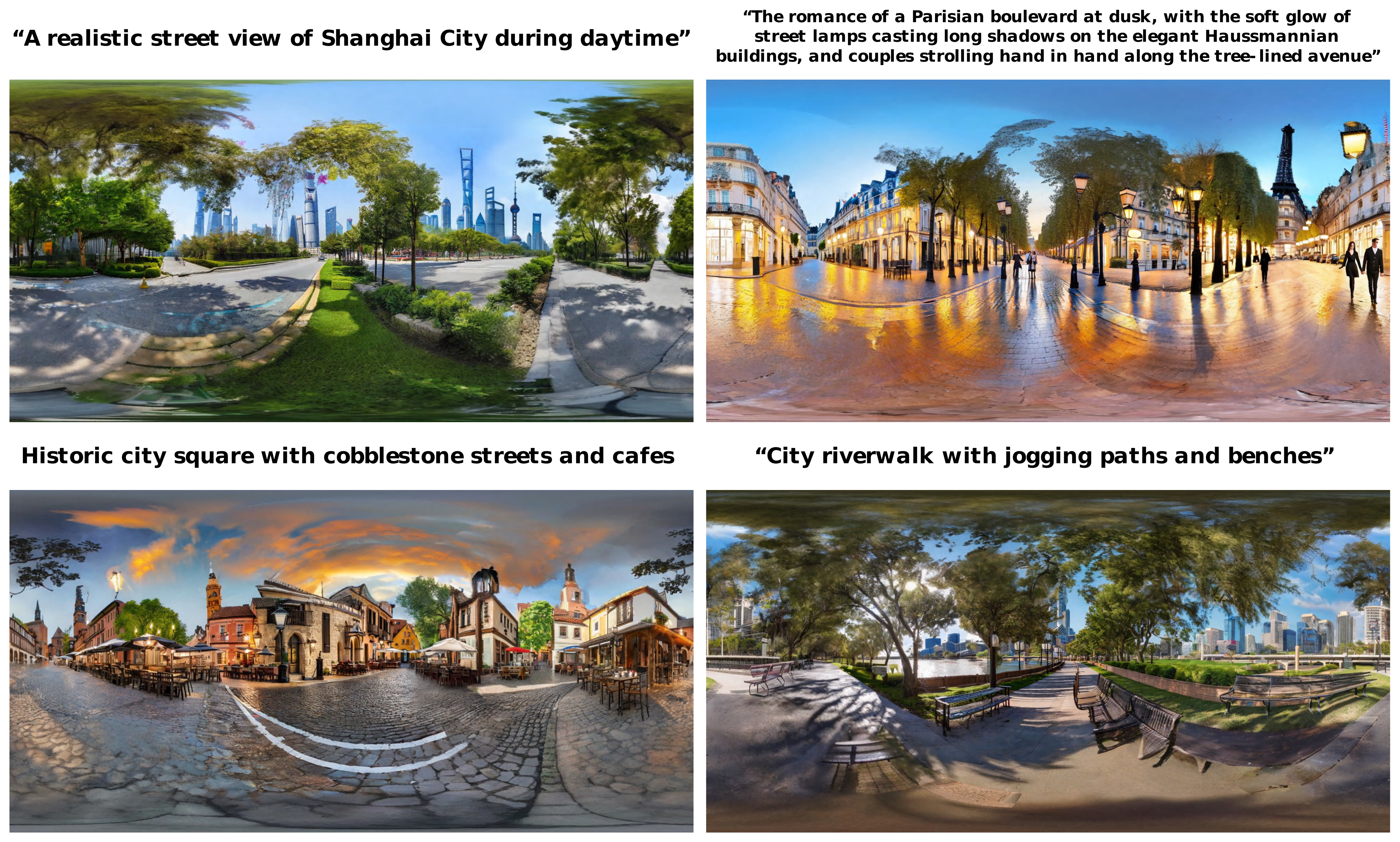}
  \vspace{-2mm}
  \caption{City and street scene full spherical panoramas generated by PanoFree.}
  \label{fig:pf_full_city}
  \vspace{-2mm}
\end{figure*}

\noindent \textbf{Natural Scene} panoramas generated with PanoFree are shown in Fig.~\ref{fig:pf_full_city}.

\begin{figure*}[h]
  \centering
  \includegraphics[width=0.99\linewidth]{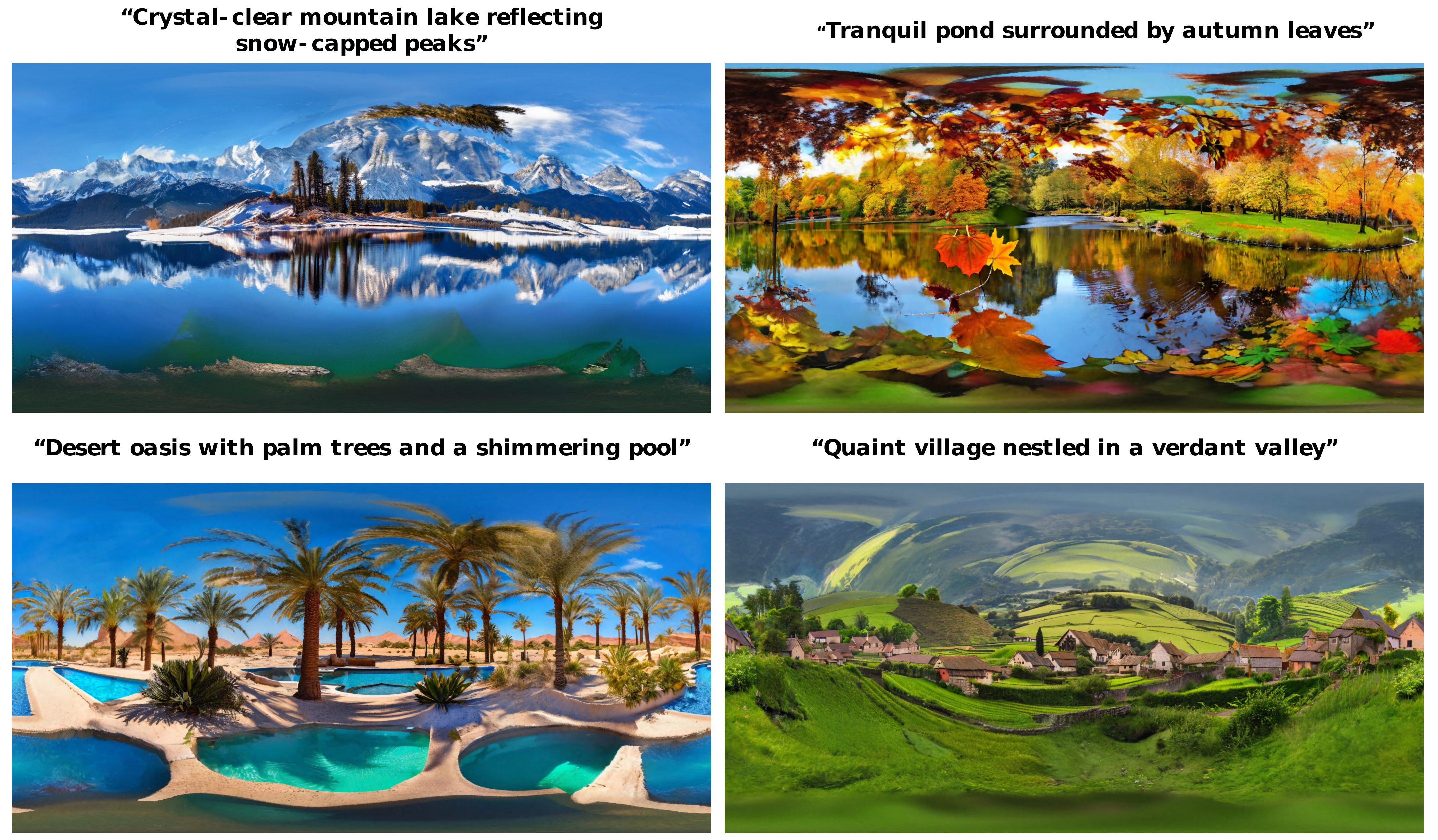}
  \vspace{-2mm}
  \caption{Natural scene full spherical panoramas generated by PanoFree.}
  \label{fig:pf_full_natural}
  \vspace{-2mm}
\end{figure*}

\noindent \textbf{Natural Scene} panoramas generated with PanoFree are shown in Fig.~\ref{fig:pf_full_natural}.

\subsection{Different Pre-trained T2I Model}
\begin{figure*}[h]
  \centering
  \includegraphics[width=0.99\linewidth]{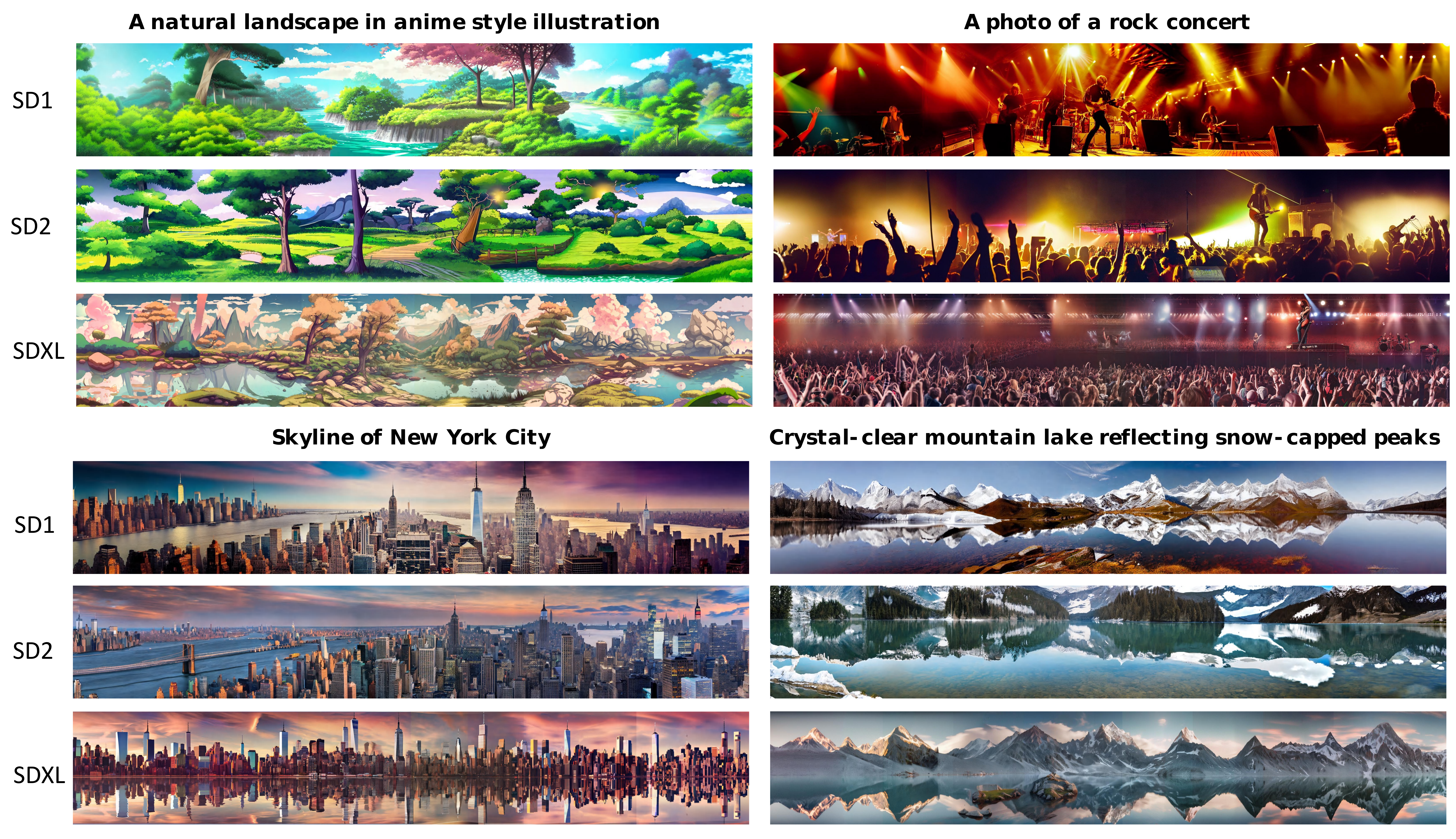}
  \vspace{-2mm}
  \caption{Panoramas generated by PanoFree using different pre-trained T2I models.}
  \label{fig:different_sd}
  \vspace{-2mm}
\end{figure*}

Additionally, PanoFree is also highly flexible and can plug-and-play with different pre-trained T2I models. In Fig.~\ref{fig:different_sd}, we illustrate this by applying Stable Diffusion v1 (SD1), Stable Diffusion v2 (SD2) and Stable Diffusion XL (SDXL) for PanoFree. The results show that PanoFree can work properly with different pre-trained T2I models.

\section{Panorama Diversity Comparison} 
In this section, we will provide more comparison results on panorama diversity and more detailed illustrations of the diversity issue of Joint Diffusion baselines~\cite{bar2023multidiffusion, lee2024syncdiffusion}.
\subsection{Planar Panorama Generation}
\begin{figure*}[h]
  \centering
  \includegraphics[width=0.99\linewidth]{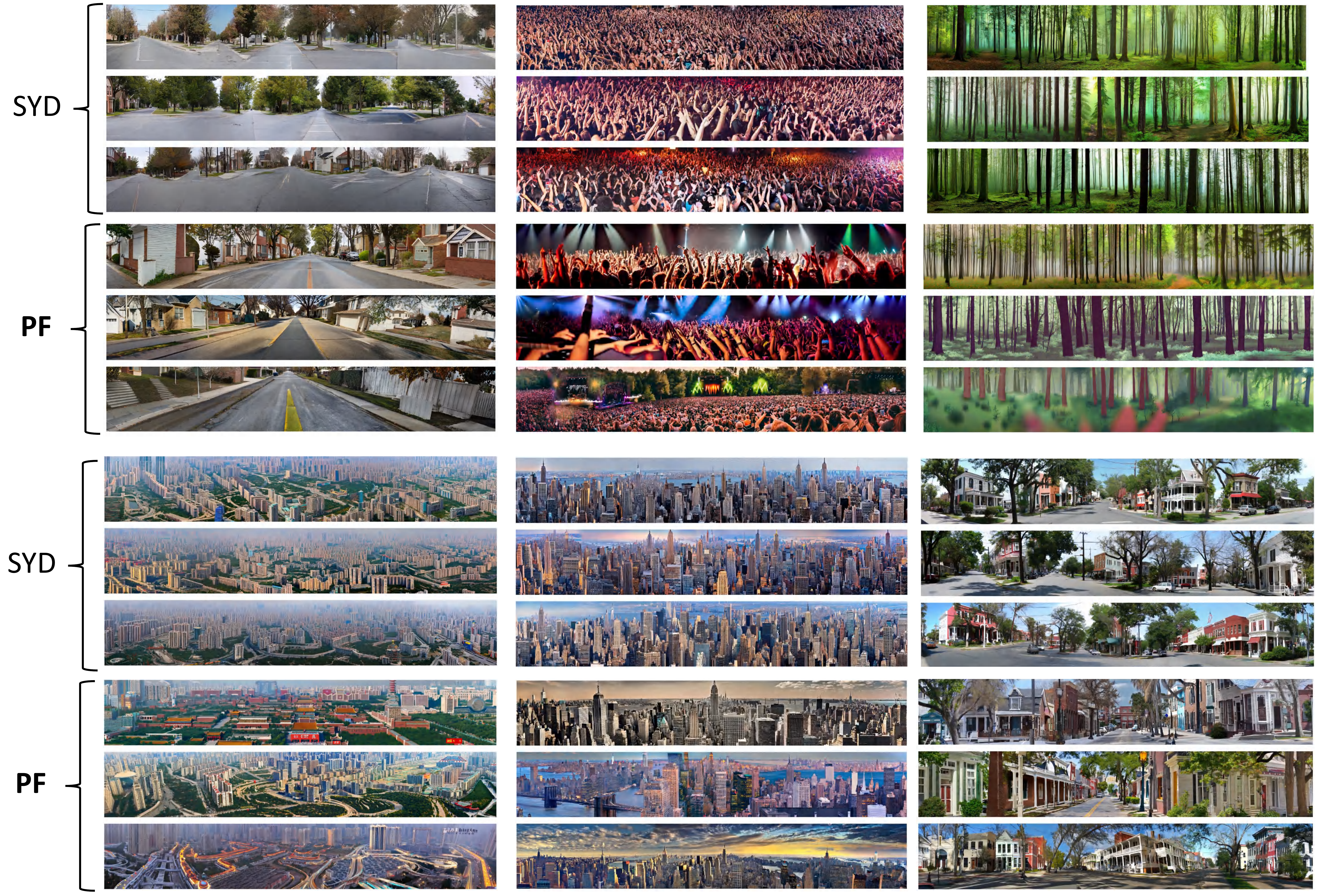}
  \vspace{-2mm}
  \caption{Planar panorama diversity illustration and comparison. PF is our PanoFree.}
  \label{fig:pf_div_planar}
  \vspace{-2mm}
\end{figure*}

For the planar panorama diversity comparison, we mainly compare PanoFree with SyncDiffusion~\cite{lee2024syncdiffusion}. For each text prompt, we select three different random seeds to generate three results. The comparison results are shown in Fig~\ref{fig:pf_div_planar}. 
We can observe that SyncDiffusion generates styles, contents, and scene structures that are very similar across different random seeds. In contrast, PanoFree can generate more diverse panoramas.
\subsection{360° Panorama Generation}
\begin{figure*}[h]
  \centering
  \includegraphics[width=0.99\linewidth]{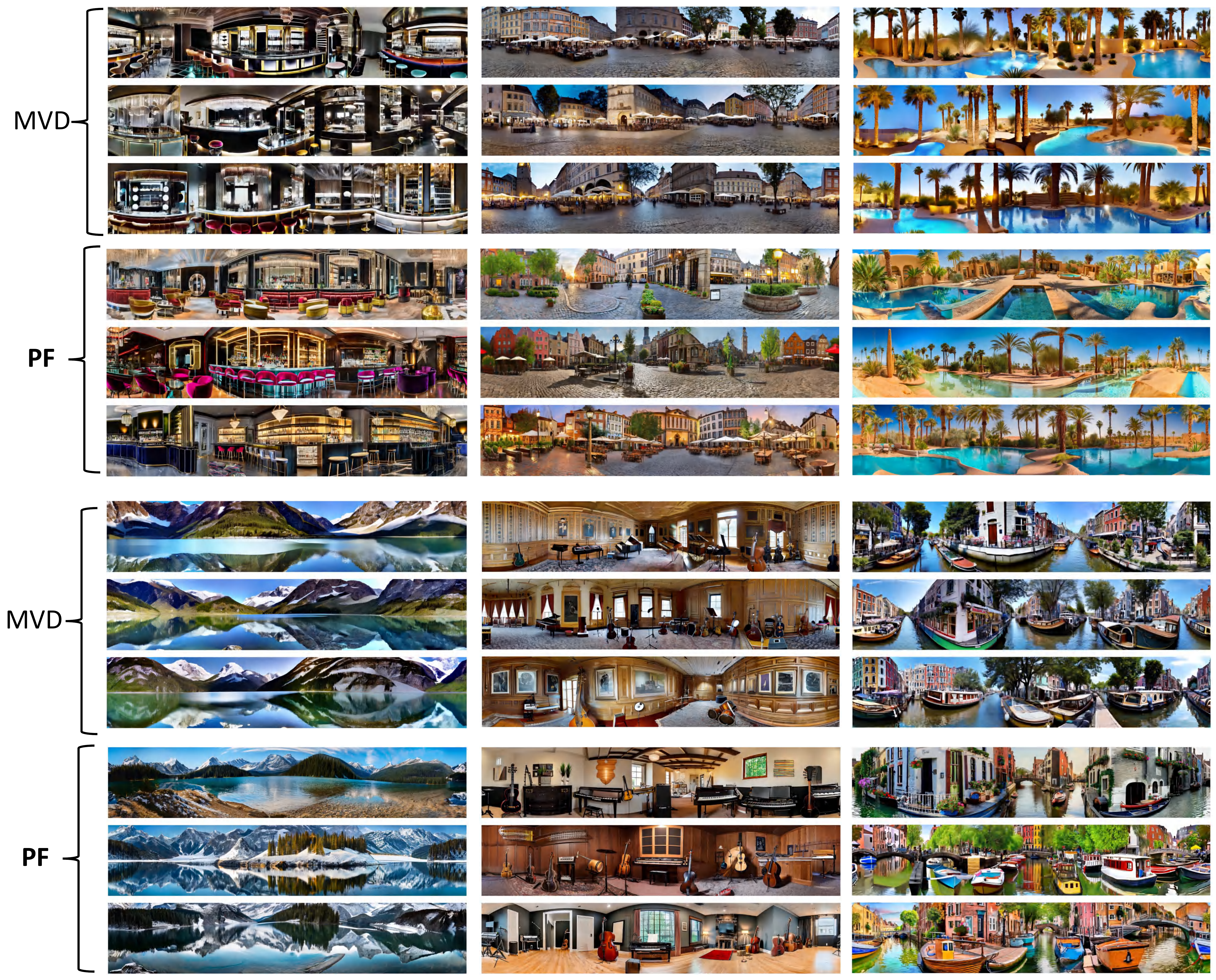}
  \vspace{-2mm}
  \caption{Planar panorama diversity illustration and comparison. PF is our PanoFree.}
  \label{fig:pf_div_360}
  \vspace{-2mm}
\end{figure*}

For the 360° panorama diversity comparison, we mainly compare PanoFree with MVDiffusion~\cite{Tang2023MVDiffusionEH}. For each text prompt, we also select three different random seeds to generate three results. The comparison results are shown in Fig~\ref{fig:pf_div_360}. We can see that MVDiffusion generates styles, contents, scene structures, etc., that are very similar across different random seeds. Even the diversity of MVDiffusion is poorer than SyncDiffusion. This is because MVDiffusion undergoes fine-tuning on top of Joint Diffusion design, which can bias the generation results towards the training dataset. In contrast, PanoFree can still generate more diverse 360 panoramas.

\subsection{Generation with Rough Prompts}

This diversity issue becomes particularly apparent when given some rough prompts. Therefore, we additionally generated a ``rough set'' consisting of 20 short and blurry prompts to exacerbate this issue. For each prompt, we used 20 different random seeds. Then, we calculated intra-LPIPS and cross-LPIPS based on the generated results from the rough set. By subtracting the intra-LPIPS from the cross-LPIPS, we can illustrate the trade-off between consistency and diversity. 
As shown in Table~\ref{table:rough_consistency_diversity}, we can observe that both MultiDiffusion and SyncDiffusion result in a significant decrease in diversity. Specifically, MultiDiffusion appears to perform an ``equivalent exchange'' between consistency and diversity.

\begin{table}[t]
\centering
\caption{Results on Planar Panorama generation with rough prompts. Intra-LPIPS ($10^{-2}$) measures global consistency, Cross-LPIPS ($10^{-2}$) diversity, and CL-IL (Cross-LPIPS - Intra-LPIPS, ($10^{-2}$)) the trade-off between consistency and diversity. }
\vspace{-2mm}
\begin{tabular}{cccc}
\toprule
 Method & Intra-LPIPS$\downarrow$  & Cross-LPIPS$\uparrow$ & CL-IL$\uparrow$ \\
\midrule
SG  & $70.67$ & $73.65$ & $2.97$  \\
MD~\cite{bar2023multidiffusion} & $68.03$ & $68.57$ & $0.54$  \\
SYD~\cite{lee2024syncdiffusion} & $64.32$ & $67.48$ & $3.16$ \\
\midrule
\textbf{PF (ours)} & $65.39$ & $72.29$& $6.90$ \\
\bottomrule
\end{tabular}
\label{table:rough_consistency_diversity}
\vspace{-4mm}
\end{table}

Note that this loss of diversity appears to be ``within the prompt''. When provided with more detailed prompts, Joint Diffusion still has the ability to generate corresponding results. However, iteratively adjusting the prompt based on the output results also incurs a considerable additional time overhead.
\section{Additional Experiment Details}
\subsection{Quantitative Evaluation Details} 
\noindent\textbf{Reference Sets.} For metrics that require a reference set, such as FID~\cite{Heusel2017GANsTB} and KID~\cite{Binkowski2018DemystifyingMG}, we generate reference sets composed of perspective view images using the same stable diffusion model with identical prompts but different random seeds. We then crop an equal number of perspective view images from the panoramas generated by PanoFree and baseline methods to perform the calculations.

\subsection{User Study Details}
For planar panorama generation and $360^{\circ}$ panorama generation, we conducted four user studies for each task to further evaluate the global consistency, image quality, prompt compatibility, and diversity of the generated panoramas, respectively. For the first three user studies, we follow the design of SyncDiffusion~\cite{lee2024syncdiffusion}. Participants were presented with panorama images generated by 2 methods and asked to measure their panorama consistency quality, prompt compatibility, and diversity (see supplementary for details). 
Then they are asked to choose one of them by answering the question: \textit{``Which one appears a more consistent panorama image to you?''} \textbf{(Consistency)}, \textit{``Which one is of higher quality?''} \textbf{(Quality)} and \textit{``Which one best matches the shared caption?''} \textbf{(Prompt Compatibility)}. 

\noindent\textbf{Diversity.} For the last user study, participants were presented with 2 groups of panoramas generated by the 2 methods in every question. Each group contains 3 panorama generated with same prompt and different random seeds. Then they are asked to choose one group by answering the question: \textit{``Which group of panorama images is more diverse''} \textbf{(Diversity)}. We set 15 questions for each user study and collect responses from 5 Amazon MTurk workers for each question.

\noindent\textbf{Baseline Methods.} For the planar panorama generation task, we selected SyncDiffusion as the primary baseline. For the 360 panorama generation task, we chose MVDiffusion as the main comparative baseline.

\subsection{Planar Panorama Generation}

\textbf{Task Configurations.} In this paper, the resolution of the generated planar panoramas is 512x3072, while each view image has a resolution of 512x512. All methods employed 50 diffusion steps. \\

\noindent\textbf{Baseline and Configuration} details are illustrated in the following.
\begin{itemize}
  \item {\it Sequential Generation (SG)} refers to the vanilla iterative warping and inpainting process. We utilize 10 warping and inpainting steps to extend the initial view image into the desired panorama, with a translation step size of 256 pixels in the image space for each step.
  \item {\it MultiDiffusion (MD)}~\cite{bar2023multidiffusion} adopts joint diffusion approach with multiple overlapping windows on the latent space. Each window has a separate diffusion process that are fused by averaging the latent features within the overlapping regions at every reverse diffusion step. We utilized default configurations from the official implementation, including a stride of 8 in the latent space.
  \item {\it SyncDiffusion (SYD)}~\cite{lee2024syncdiffusion} is another joint diffusion approach which achieves the state-of-the-art in Planar Panorama generation regarding global consistency. SyncDiffusion fuses multiple diffusion process and ensures global consistency by guiding the reverse diffusion process while adjusting the intermediate latent features at every step. We used default configurations from the official implementation, including a stride of 16 in the latent space, a weight of 20, and a weight decay with a rate of 0.95.
\end{itemize}

\noindent\textbf{PanoFree Configurations.} Similar to vanilla sequential generation, PanoFree also utilizes 10 warping and inpainting steps, 5 steps in each direction, to extend the initial view image into the desired panorama, with a translation step size of 256 pixels in the image space for each step. For SDEdit, we set $t_0 = 0.98$. We only estimate risk based on the distance to the initial view. At each step, we erase 30\% of known areas based on the estimated risk.

\subsection{360° Panorama Generation}
\textbf{Task Configurations.} 
In this paper, the spherical surface is represented by a 2048x4096 2D image with equirectangular projection and we care about area with pitch $\in [-40, 40]$ for 360° panoramas. Each view image has a resolution of 512x512. All methods employed 50 diffusion steps. 

\noindent\textbf{Baselines.} We have chosen 2 baselines for comparison. Except the {\it Vanilla Sequential Generation (SG)}, we also selected {\it MVDiffusion (MVD)} as a baseline. The baseline details and implementation details are available in the appendix.
\begin{itemize}
  \item {\it Sequential Generation (SG)} still refers to the vanilla iterative warping and inpainting process. The only difference is that the current warping corresponds to optical geometric changes caused by rotation. We adopt a Field of View (FoV) of 80° and a yaw stride of 40°. 8 warping and inpainting steps were utilized.
  \item {\it MVDiffusion (MVD)}~\cite{Tang2023MVDiffusionEH} also adopts the Joint Diffusion design, which fuses multiple diffusion processes together to generate consistency in multi-view images by incorporating correspondence-aware attention into a pretrained diffusion model. MVDiffusion requires panorama images for training, and we chose to utilize the model weights provided by the authors. It is worth noting that although the model weights were trained on indoor scenes, MVDiffusion demonstrates impressive generalization ability and can generate outdoor data. We utilized the default configurations from the official implementation.
\end{itemize}
\textbf{PanoFree Configurations.} Similar to vanilla sequential generation, PanoFree also utilizes a Field of View (FoV) of 80° and a yaw stride of 40°. 7 warping and inpainting steps were utilized, including 3 symmetric steps in each direction, and a merging step. For SDEdit, we set $t_0 = 0.98$. We use $\mathbf{w} = [0.8, 0.2, 0, 0]$ to combine the risks $[\mathbf{r}^{i}, \mathbf{r}^{e},\mathbf{r}^{c}, \mathbf{r}^{s}]$. At each step, we erase 5\% of known areas based on the estimated risk.
\subsection{Full Spherical Generation}

\textbf{Task Configurations.} 
In this paper, the spherical surface is represented by a 2048x4096 2D image with equirectangular projection and we care about the whole spherical surface for full spherical panoramas. Each view image has a resolution of 512x512. All methods employed 50 diffusion steps. 

\noindent\textbf{PanoFree Configurations.} The configurations for generating areas with pitch $\in$ [-40°, 40°] are exactly the same as described for the 360° panorama. Then, we rotate the viewpoint upwards and downwards by 25° to generate areas with pitch $\in$ [40°, 65°] and pitch $\in$ [-40°, -65°]. During expansion, we use a Field of View (FoV) of 110° and a stride of 80°. Three warping and inpainting steps are required for expansion in both the upward and downward directions. Finally, using the upper and lower poles as centers, we use one warping and inpainting step each to generate areas with pitch $\in$ [65°, 90°] and pitch $\in$ [-65°, -90°]. For expansion stages, we set $t_0 = 0.90$. We use $\mathbf{w} = [0.6, 0.2, 0.1, 0.1]$ to combine the risks $[\mathbf{r}^{i}, \mathbf{r}^{e},\mathbf{r}^{c}, \mathbf{r}^{s}]$. The guidance scale is set to 2.0, while the variance of noise is amplified by a factor of 1.05. At each expansion step, we erase 10\% of known areas based on the estimated risk. At the final close-up stpes, we use a Field of View (FoV) of 90°. We set $t_0 = 0.90$. We use $\mathbf{w} = [0.6, 0.2, 0.1, 0.1]$. And we erase 20\% of known areas based on the estimated risk. The guidance scale is set to 1.0, while the variance of noise is amplified by a factor of 1.1.

\section{Limitation and Failure Cases}

In this section, we discuss about some limitations and failure cases of PanoFree.

\subsection{Undesired Camera Pose} 
\begin{figure}[h]
  \centering
  \begin{subfigure}{0.95\linewidth}
    \includegraphics[width=0.95\linewidth]{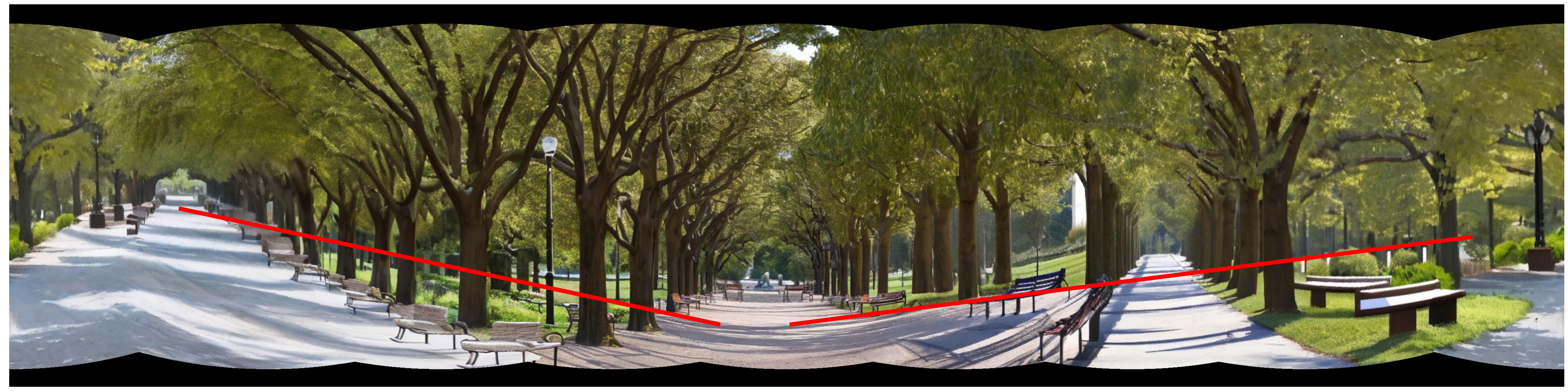}
    \caption{Undesired camera poses}
    \label{fig:ill_cam}
  \end{subfigure}
  \\
  \begin{subfigure}{0.95\linewidth}
    \includegraphics[width=0.95\linewidth]{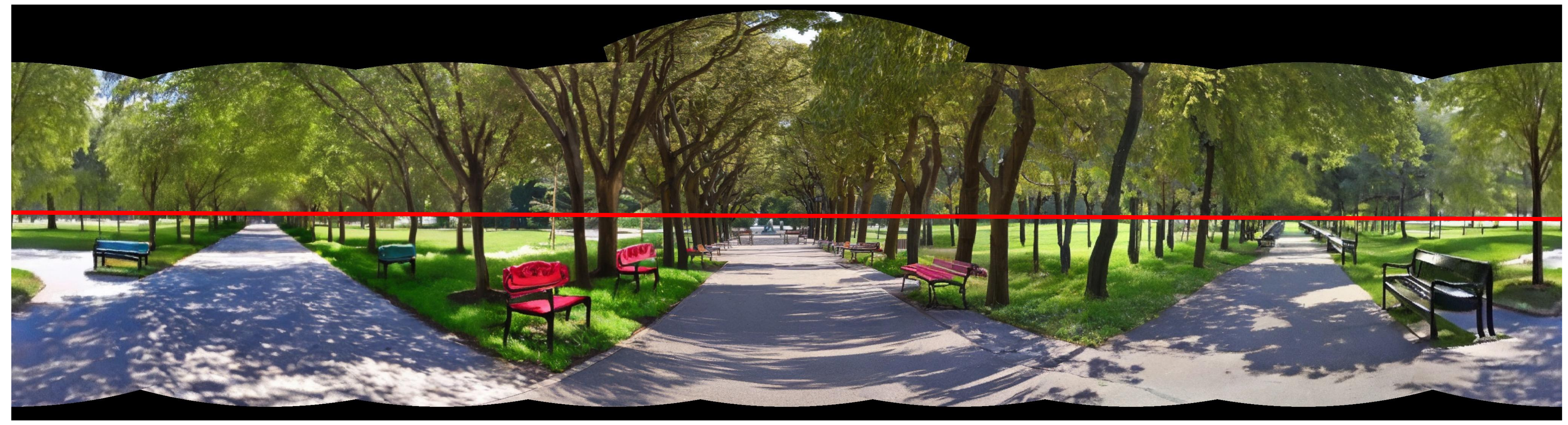}
    \caption{Estimate the camera pose of initial view}
    \label{fig:init_cam}
  \end{subfigure}
  
  \vspace{-2mm}
  \caption{Results with the undesired camera poses and the estimated initial camera pose.}
  \label{fig:cam_poses}
    \vspace{-4mm}
\end{figure}

refers to the inconsistency between the underlying camera pose of the generated images and the camera pose we set. Undesired camera pose issue can lead to failure cases when generating 360 panoramas. As illustrated in Fig.~\ref{fig:ill_cam}, this issue often results in ground deformation and may cause severe distortion, thereby making the generated panorama appear unreasonable. 

Potential solutions include fine-tuning the T2I models with images having desired camera poses or incorporating the camera pose as an additional model conditioning input to control the generated images. However, these approaches require costly fine-tuning. In this paper, we narrow down this problem to the camera pose of the initial view: we find that as long as the underlying camera pose of the initial view image is relatively close to our set pose, there are fewer occurrences of undesired camera pose issues in the following views. Therefore, as illustrated in Fig~\ref{fig:init_cam}, we can mitigate this problem by leveraging open-source pre-trained camera pose estimation models to predict the camera pose of the initial view image. Specifically, we primarily care about the pitch and Field of View (FoV) of initial view image.

\subsection{Biased Generation}

\begin{figure*}[h]
  \centering
  \includegraphics[width=0.95\linewidth]{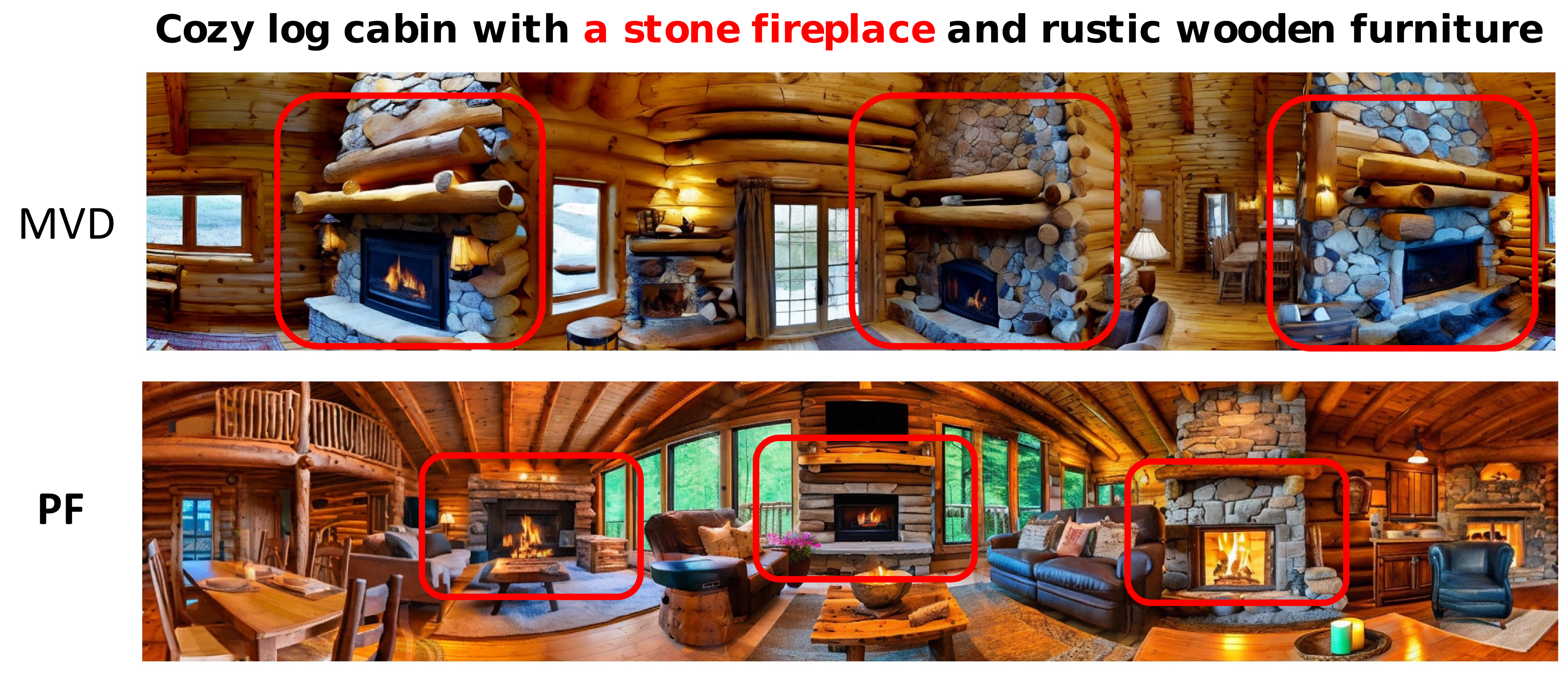}
  \vspace{-2mm}
  \caption{Duplicated semantic contents generated with MVDiffusion~\cite{Tang2023MVDiffusionEH} and PanoFree.}
  \label{fig:semantic_dup}
  \vspace{-2mm}
\end{figure*}

\begin{figure}[h]
    \vspace{-2mm}
  \centering
  \begin{subfigure}{0.95\linewidth}
    \centering
    \includegraphics[width=0.99\linewidth]{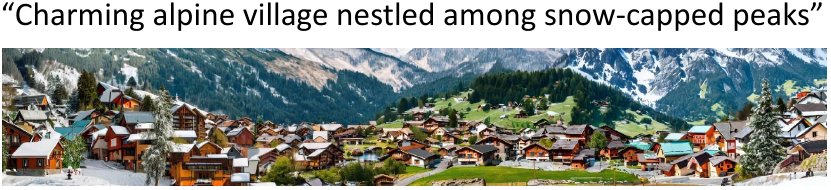}
    \caption{Biased Generation.}
    \label{fig:biased_generate}
  \end{subfigure}

  \begin{subfigure}{0.95\linewidth}
    \centering
    \includegraphics[width=0.99\linewidth]{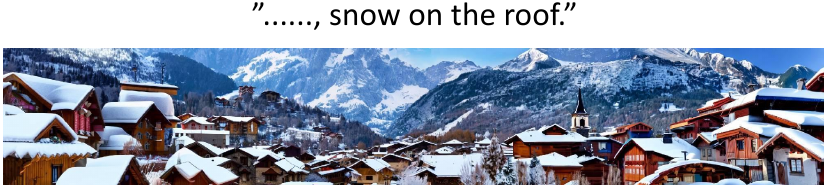}
    \caption{Correction with dense prompt.}
    \label{fig:dense_prompt}
  \end{subfigure}
  
  \vspace{-2mm}
  \caption{Correcting biased generation with denser prompts.}
  \label{fig:bias_correction}
  \vspace{-4mm}
\end{figure}

Another issue is biased generation PanoFree may exhibit a bias towards ensuring local alignment with the prompt and scene priors, potentially leading to conflicts on a global scale. There are two common issues. Firstly, as shown in Fig~\ref{fig:semantic_dup}, PanoFree may to generate duplicated semantic contents across the panorama. Secondly,, it can produce inconsistent scene characteristics in different parts of the image. For instance, in the left image of Fig.~\ref{fig:biased_generate},one section depict a winter landscape, while another section simultaneously presents spring-like features.

However, it's important to note that PanoFree's primary contribution lies in its tuning-free and efficient panorama generation approach, rather than completely eliminating these biases. On the other hand, biased generation remains a challenge in panorama generation tasks, affecting many methods including those trained on real panoramic data. For example, As shown in Fig~\ref{fig:semantic_dup}, MVDiffusion~\cite{Tang2023MVDiffusionEH} also exhibits issues with duplicated semantic content.  Nevertheless, if data and computational costs are not constraints, PanoFree can be readily enhanced. For example, pretrained LLMs could be employed to generate denser prompts, potentially correcting biased generation, as demonstrated in Fig.~\ref{fig:dense_prompt}.

\end{document}